\documentclass[journal]{IEEEtran}

\usepackage{graphicx}
\usepackage{booktabs}
\usepackage{multirow}
\usepackage{adjustbox}
\usepackage{amsmath}
\usepackage{amssymb}
\usepackage{pifont}
\usepackage{stackengine}

\usepackage{caption}
\usepackage{subcaption}
\usepackage{array}
\usepackage{cite}
\usepackage[dvipsnames]{xcolor}
\usepackage{hyperref}
\usepackage{times}
\usepackage{epsfig}
\usepackage{tablefootnote}
\usepackage{tabularx}

\usepackage{arydshln}

\definecolor{diagram_green}{RGB}{130, 179, 102}
\definecolor{diagram_yellow}{RGB}{214, 182, 86}
\definecolor{diagram_red}{RGB}{184, 84, 80}
\definecolor{diagram_blue}{RGB}{108, 142, 191}
\definecolor{diagram_orange}{RGB}{215, 155, 0}
\definecolor{teaser_blue}{RGB}{37, 150, 190}

\usepackage{bm}

\newcommand{\red}[1]{\textcolor{red}{#1}}
\newcommand{\blue}[1]{\textcolor{blue}{#1}}
\newcommand{\gray}[1]{\textcolor{gray}{#1}}

\newcommand{\textbfred}[1]{\textcolor{red}{{#1}}}

\usepackage[capitalize]{cleveref}
\crefname{section}{Sec.}{Secs.}
\Crefname{section}{Section}{Sections}
\Crefname{table}{Table}{Tables}
\crefname{table}{Tab.}{Tabs.}

\begin{document}

\title{DisCoVQA: Temporal Distortion-Content Transformers for Video Quality Assessment}

\author{Haoning Wu,
        Chaofeng Chen,
        Liang Liao,~\IEEEmembership{Member,~IEEE},
        Jingwen Hou,~\IEEEmembership{Student Member,~IEEE},\\
        Wenxiu Sun,
        Qiong Yan,
        Weisi Lin,~\IEEEmembership{Fellow,~IEEE}
\thanks{H. Wu, C. Chen, L. Liao, J. Hou, and W. Lin are with the School of Computer Engineering, Nanyang Technological University, Singapore. (e-mail: haoning001@e.ntu.edu.sg; chaofeng.chen@ntu.edu.sg; liang.liao@ntu.edu.sg; jingwen003@e.ntu.edu.sg; wslin@ntu.edu.sg;)}
\thanks{W. Sun and Q. Yan are with the Sensetime Research, Hong Kong, China. (e-mail: irene.wenxiu.sun@gmail.com;  sophie.yanqiong@gmail.com)}
\thanks{Corresponding author: Weisi Lin.}}

\markboth{IEEE Transactions on XX,~Vol.~XX, No.~X, MONTH~YEAR}%
{Shell \MakeLowercase{\textit{et al.}}: Bare Demo of IEEEtran.cls for IEEE Journals}

\maketitle

\begin{abstract}

\textcolor{black}{The temporal relationships between frames and their influences on video quality assessment (VQA) are still under-studied in existing works. These relationships lead to two important types of effects for video quality. Firstly, some temporal variations (such as shaking, flicker, and abrupt scene transitions) are causing temporal distortions and lead to extra quality degradations, while other variations (e.g. those related to meaningful happenings) do not. Secondly, the human visual system often has different attention to frames with different contents, resulting in their different importance to the overall video quality. Based on prominent time-series modeling ability of transformers, we propose a novel and effective transformer-based VQA method to tackle these two 
issues. To better differentiate temporal variations and thus capture the temporal distortions, we design a transformer-based Spatial-Temporal Distortion Extraction (STDE) module. To tackle with temporal quality attention, we propose the encoder-decoder-like temporal content transformer (TCT). We also introduce the temporal sampling on features to reduce the input length for the TCT, so as to improve the learning effectiveness and efficiency of this module. Consisting of the STDE and the TCT, the proposed Temporal Distortion-Content Transformers for Video Quality Assessment (\textit{DisCoVQA}) reaches state-of-the-art performance on several VQA benchmarks without any extra pre-training datasets and up to 10\% better generalization ability than existing methods. We also conduct extensive ablation experiments to prove the effectiveness of each part in our proposed model, and provide visualizations to prove that the proposed modules achieve our intention on 
modeling these temporal issues. We will publish our codes and pretrained weights later.}

\end{abstract}

\begin{IEEEkeywords}
Deep learning, video quality assessment, temporal modeling, transformers
\end{IEEEkeywords}

%
\IEEEpeerreviewmaketitle

\section{Introduction} \label{sec:intro}

\IEEEPARstart{W}{ith} the rapid development of smartphones and portable cameras, more and more videos are created every day by a huge diversity of consumers, both professional or non-professional users.
These videos are collected in-the-wild and uploaded to popular social media websites or apps such as YouTube and TikTok, and the number of them is still growing. Henceforth, it becomes increasingly important to build a robust and powerful objective VQA method for these natural videos without pristine references.




Many existing VQA efforts \cite{vsfa,tlvqm,vbliinds,videval,rfugc,pvq} have demonstrated that directly adopting the image quality assessment (IQA) models to measure video quality with frame-wise IQA procedures is not effective for VQA, as they neglect the temporal relationships between frames. Specifically, TLVQM\cite{tlvqm} have demonstrated that the variations between frames, such as shaking, flicker and abrupt scene transitions, are causing additional quality degradations. Such effects have also been observed by many deep VQA approaches such as CoINVQ\cite{rfugc} and PVQ\cite{pvq}. We categorize all these degradations caused by variations between different frames as \textit{\textbf{temporal distortions}}, the first type of temporal issues in VQA. While these distortions are usually noticed, many approaches at present still apply handcrafted kernels to model them. Despite temporal distortions, many existing approaches have also noticed that different frames should have different importance to the final video quality. VSFA\cite{vsfa} and some other recent approaches \cite{mdtvsfa, bvqa2022} have concluded this effect as ``temporal memory effect'' that early frames are prone to be forgotten, yet recent studies on the human visual system (HVS) such as \cite{febrain} imply that importance of frames should be highly related to their contents, so we categorize this effect as \textit{\textbf{content-related temporal quality attention}} instead. These two temporal issues are often mentioned in prior arts but less systematically discussed or modeled. Recently, transformer architectures 
are proved to have better ability on time-series modeling in several tasks, including language modeling \cite{allyouneed} and video recognition \cite{vivit,mvit,swin3d}, which allows our method to revisit these two issues discussed above with powerful transformers. In the following two paragraphs, we will discuss them in detail and provide our corresponding approaches based on transformers.


\begin{figure*}
    \centering
    \includegraphics[width=1.0\linewidth]{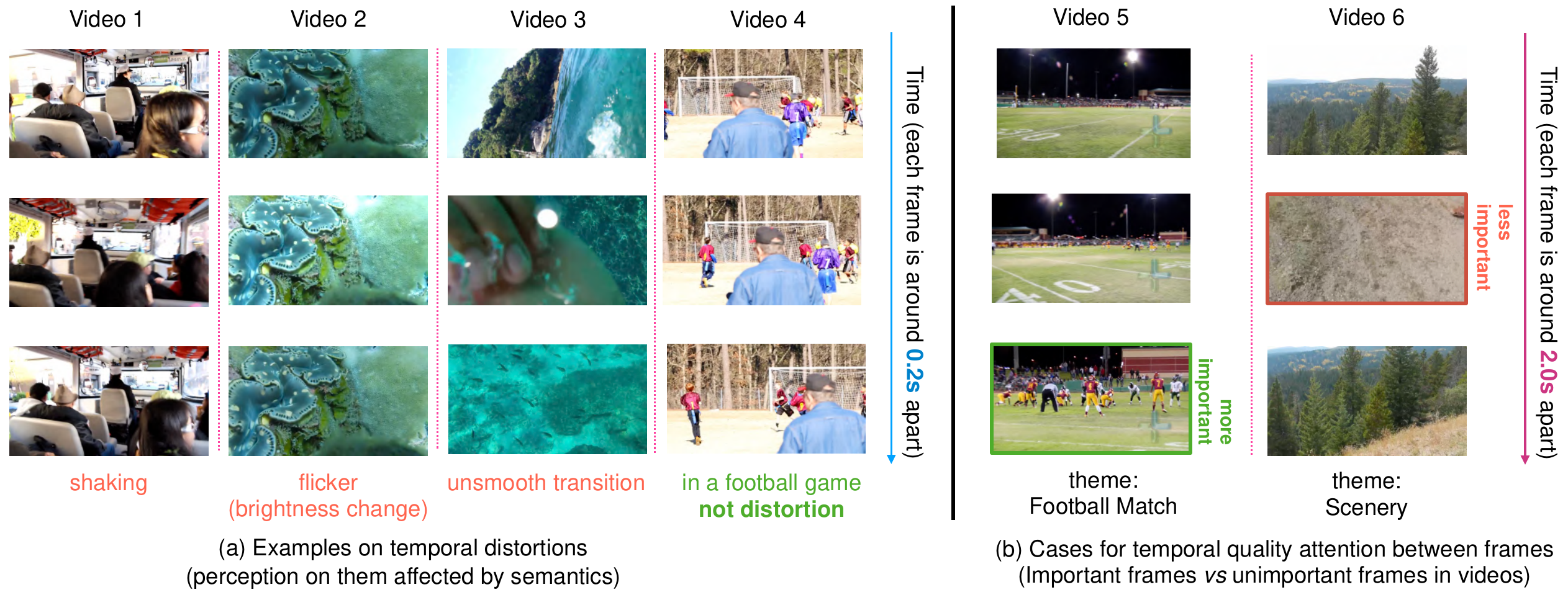}
    \caption{Examples for naturally collected videos in LSVQ dataset\cite{pvq} show complicated temporal relationships and their effects to VQA. These effects have shown that better temporal modeling are in need for VQA task.}
    \label{fig:gl}
\end{figure*}

First of all, for temporal distortions, it is straightforward to notice that they come from temporal variations between frames, as static videos without variations do not have temporal distortions. As illustrated in \cref{fig:gl}(a), some temporal variations, such as the shaking of the whole picture in \textit{video 1}, the rapid brightness changes among frames in \textit{video 2} and the abrupt transition of the scene in \textit{video 3}, do not have clear semantics or reason and lead to temporal distortions.  
On the other hand, not all temporal variations lead to temporal distortions. To be specific, some other temporal variations come from human actions or other meaningful activities/events in videos, such as running players and correspondingly moving scene during a football game (\textit{video 4}), and therefore do not lead to distortions in general. 
The aforementioned two scenarios suggest that understanding the semantics of actions (or other activities) helps to better capture temporal distortions. Therefore, we adopt the Video Swin Transformer Tiny (\textit{abbr.} as Swin-T) with better action recognition ability as the backbone for our Spatial-Temporal Distortion Extraction (STDE). Such semantic information cannot be captured by existing classical handcraft models \cite{tlvqm, videval, vbliinds} and even 3D-CNNs \cite{k400, r3d, c3d} as introduced in \cite{cnn+lstm, pvq}, which are less effective than transformers on several action-related tasks as compared in \cite{timesformer,mvit,swin3d}. In addition, we apply the temporal differences to further enhance the sensitivity for temporal variations, especially for those variations independent to actions (\textit{e.g.} scene transitions) and are less captured by the backbone network. With Swin-T and temporal differences, we better extract features that are sensitive to distortion-related temporal variations in STDE.



Besides temporal distortions, the temporal quality attention mechanism also affects the overall quality evaluation of the video. Similar to spatial quality attention that is related to contents of different regions \cite{musiq,hyperiqa}, the temporal attention should also be related to contents of frames. For example, as shown in \cref{fig:gl}(b), \textit{video 5}, in the replay video of a football match, the attention of the HVS is more attracted by the frames that contain the zoomed-in players (in the green box), which are closer to what the whole video is about, \textit{i.e.} the video theme. On the contrary, some frames less related to the video theme might be less important in deciding the final video quality. For example, the poor quality of the intermediate frames (in the red box) shot to the ground does not affect the overall video quality of \textit{video 6} rated as good (as scored by the human in \cite{pvq}), as the HVS pays less attention on such frames but more to those frames about \textit{Scenery} (theme of \textit{video 6}). Both examples suggest that the relevance of frame contents 
to the overall video theme affects the importance of frames in deciding the final video quality. As transformers are well recognized as good at learning correlations among a sequence \cite{bert}, \cite{allyouneed}, we propose the temporal content transformer (TCT) to learn the correlations of frame contents and model this temporal quality attention. The TCT has an encoder-decoder-like transformer structure that better extracts the temporal quality attention from frame contents. Some existing approaches \cite{vsfa, cnn+lstm, mdtvsfa, gstvqa} also introduce RNNs such as GRU \cite{gru} or LSTM \cite{lstm} to model the temporal quality attention. However, RNN-based models are usually weak in modeling long-range correlation, so they are less effective in extracting the correlations between frames across the whole video and their relevance to the overall theme, which is better modeled by the proposed TCT.

\begin{figure*}[t]
    \centering
    \includegraphics[width=\linewidth]{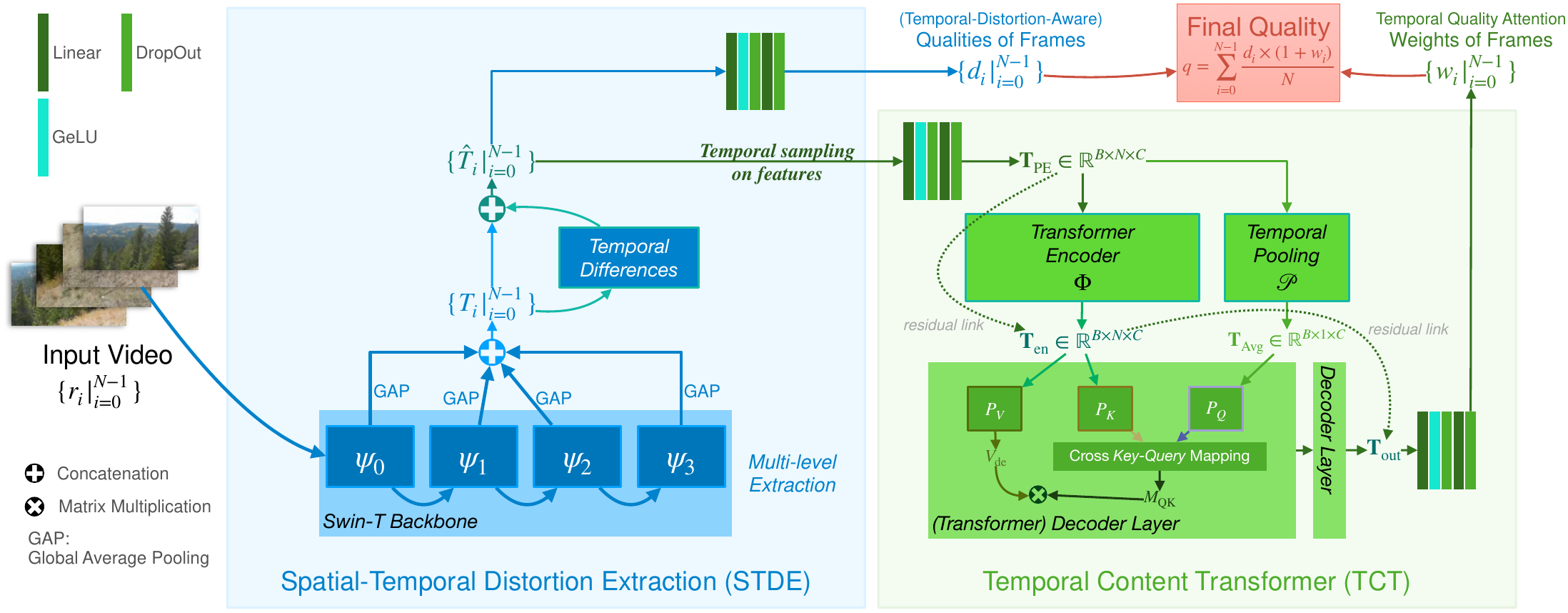}
    \caption{The structure of proposed Temporal Distortion-Content Transformers for Video Quality Assessment (\textbf{DisCoVQA}). It contains the Spatial-Temporal Distortion Extraction (STDE, \cref{sec:31}) to better extract temporal distortions (such as \textit{shaking, flicker, unsmooth transitions}), and the Temporal Content Transformer (TCT, \cref{sec:32}) to learn the content-related temporal quality attention between frames.}
    \label{fig:DisCoVQA}
\end{figure*}

As analyzed above, we introduced two transformer-based modules, the STDE and the TCT, to improve temporal modeling in VQA. To better apply transformers in VQA, we present two vital designs to improve the effectiveness of the two transformer-based modules. For the STDE, as the last-layer output of the transformer backbone 
might be less sensitive to low-level features and temporal variations, we introduce multi-level feature extraction to capture both low-level and high-level features and thus better spot temporal distortions. For the TCT, the transformer-based architecture is hard to be learnt effectively when datasets are small or input sequences are too long. Also, the long inputs lead to quadratically-increased computational costs. To improve the training effectiveness and efficiency, we propose the temporal sampling on features (TSF) to reduce the input length for the TCT during training. By cutting video features into segments and randomly sampling one feature frame from each segment, the TSF significantly improves the performance of training the TCT on small or long-duration video datasets. With the STDE, the TCT, and the vital designs discussed above, we build an efficient and effective transformer-based VQA model, as illustrated in \cref{fig:DisCoVQA}.

 In summary, we discuss the temporal relationships in VQA and propose the \textbf{T}emporal \textbf{Dis}tortion-\textbf{Co}ntent Transformers for \textbf{V}ideo \textbf{Q}uality \textbf{A}ssessment~(\textbf{DisCoVQA}) that reaches state-of-the-art performance and good generalization ability on most natural VQA datasets, with the following contributions.

\begin{itemize} 
    \item We propose the Spatial-Temporal Distortion Extraction (STDE) to model the temporal distortions with consideration of the impact of semantics of actions in videos. It includes a Swin-T backbone and computes temporal differences to better model distortion-related temporal variations during the feature extraction phase in VQA.
    \item We propose the Temporal Content Transformer (TCT) to learn the temporal quality attention, considering the correlation of frames with the overall video theme. It includes a transformer encoder and a transformer decoder to learn quality attention weights for frames. 
    \item We propose two important designs to adapt transformers to VQA: multi-level extraction to enhance low-level sensitivity for the STDE; and temporal sampling on features (TSF) to reduce the input length and improve the training effectiveness for the TCT. They both improve the final performance of the proposed DisCoVQA.
\end{itemize}

In the rest of the paper, we present the related work in Section II and the proposed temporal distortion-content transformer (DisCoVQA) in Section III. In section IV, we elaborate experimental settings and present extensive experimental results. In section V, we conduct the ablation studies, qualitative results and reliability analysis to explain its effectiveness. Finally, conclusions are drawn in Section VI.


\section{Related Works}
In this section, we briefly review related works in each of the three sub-fields: classical video quality assessment (VQA) methods, prior deep VQA methods, and transformers on temporal modeling.

\subsection{Classical VQA methods}
Classical VQA methods design on spatial \cite{brisque, friquee, hosa} or spatial-temporal \cite{vbliinds, viideo, tlvqm, videval} handcrafted features and regress on these features for quality modeling. Among them, TLVQM \cite{tlvqm} introduces a combination of spatial high-complexity and temporal low-complexity handcraft features and has reached state-of-the-art performance on LIVE-VQC \cite{vqc}. VIDEVAL \cite{videval} ensembles different handcraft features and achieves better performance on KoNViD-1k \cite{kv1k}. Several most recent models, such as RAPIQUE\cite{rapique} and CNN-TLVQM\cite{cnntlvqm}, use handcraft features for temporal modeling and deep neural networks for spatial modeling and reach far better results than pure spatial modeling approaches. Classical VQA methods have proved that the temporal relationships are non-negligible when evaluating video quality.

\subsection{Deep VQA Methods}
During recent years, deep video quality assessment (VQA) methods have become predominant in VQA. Instead of extracting handcraft features, deep VQA methods \cite{gstvqa, vsfa, mlsp} extract rich semantic features with CNN and regress the extracted features to predict video quality. For example, MLSP-FF \cite{mlsp} runs a linear regression on features extracted from video frames with Inception-ResNet-v2 \cite{irnv2}. These deep VQA methods also benefits from better temporal modeling. For instance, to model temporal distortions more effectively, several works \cite{cnn+lstm, pvq, rfugc} also introduce 3D-CNN instead of 2D-CNN to extract features. Some other deep VQA methods also use recurrent neural network (RNN) layers as temporal modeling for VQA, aiming to capture the temporal quality attention. For example, VSFA \cite{vsfa} uses ResNet-50 backbone \cite{he2016residual} and GRU \cite{gru} as temporal regression, while some others \cite{cnn+lstm, rfugc} also use LTSM \cite{lstm}, another type of RNN instead of GRU. LSCT-PHIQ\cite{lsctphiq} introduces naive transformer encoders for temporal attention modeling. A most recent model, \textit{R+S}\cite{bvqa2022}, combines CNN-based temporal distortion modeling and RNN-based temporal quality attention modeling together and reaches good performance. The existing practice of deep VQA methods has further demonstrated the importance of temporal distortion representation and temporal quality attention modeling, which can significantly improve the prediction accuracy for VQA.

\subsection{Transformers on Temporal Modeling}

Transformers proposed by \cite{allyouneed} have been proved to have a powerful self-attention mechanism on time-series modeling. Many representative works such as BERT\cite{bert} and GPT-3\cite{gpt} have shown transformers can be far better on extracting attention from long sequences than RNNs (\cite{gru,lstm}) on language tasks. Transformers have also boosted video-related vision tasks. Many  transformer-based video backbones, \textit{e.g.} ViViT\cite{vivit}, MViT\cite{mvit} and Video Swin Transformers \cite{swin3d}, have achieved even better performance on video recognition with pure videos than conventional CNN-based methods combined with optical flows and much better than pure 3D-CNN methods. As optical flows are originally designed to explicitly compute temporal variations, the success of video transformer backbones suggests that they are already strong on modeling temporal variations. All these existing practices have suggested transformers are stronger structures for different types of temporal modeling.

\section{Proposed Method}

\subsection{Framework Overview}
The proposed DisCoVQA adopts a temporal hierarchical pipeline, as shown in \cref{fig:DisCoVQA}. In DisCoVQA, we first input a video into the STDE (\cref{sec:31}). The STDE extracts feature tokens that are sensitive to temporal distortions (shaking, flicker, unsmooth transitions). Then the tokens are sampled by the temporal sampling on features (TSF) and passed into the TCT (\cref{sec:32}) to further model their content-related temporal quality attention. These attention weights are multiplied with the temporal-distortion-aware qualities of frames regressed by the STDE to obtain the overall quality prediction (\cref{sec:33}). Each of these components will be explained below.

\subsection{Spatial-Temporal Distortion Extraction} \label{sec:31}

The Spatial-Temporal Distortion Extraction (STDE) is designed to better capture inter-frame temporal distortions during the feature extraction stage of VQA. It first extracts multi-level features from the Swin-T backbone, and compute temporal differences on extracted features, and then regress these features into temporal-distortion-aware qualities of frames. We explain each part of the STDE as follows.

\subsubsection{Multi-level Feature Extraction on Swin-T Backbone} In STDE, we first extract features with a transformer-based backbone. For evaluation fairness, we choose the Video Swin Transformer Tiny (\textit{abbreviated as} Swin-T) instead of its heavier versions which has similar parameters with I3D-ResNet-50\cite{k400} (the most common ResNet-50 variant for videos) as our backbone. The Swin-T consists of four hierarchical Swin Transformer blocks ($\Psi_\text{l},l=0,1,2,3$), where each block has several alternate 3D window multihead self-attention (3DW-MSA) layers and feed-forward layers. A video clip $\mathcal{R} = \{r_i|_{i=0}^{N-1}\}$ is passed into Swin-T, where $i$ is the index for the $i$-th frame. Then the feature set $\mathcal{M}^l=\{M_i^l|_{i=0}^{N-1}\}$ for each frame of the video clip $\mathcal{R}$ is obtained from $l+1$ cascading blocks of Swin-T:
\begin{equation}
\mathcal{M}^l = \Psi_l(...(\Psi_0(\mathcal{R}))),~~~~l=0,1,2,3,
\end{equation}
Though adopting $l=3$ is enough for recognition tasks, low-level quality-related information may not be sufficiently preserved in the resulting features. Therefore, we introduce the multi-level feature extraction on the Swin-T backbone to enhance sensitivity of features on low-level distortion-related information.
During multi-level feature extraction, the features from different levels of Swin-T $\mathcal{M}^l$ are spatially pooled by the global average pooling layer ({GAP}), and concatenated into the primary tokens:
\begin{equation}
    \mathcal{T} = \bigoplus_{l=0}^{3} \text{GAP}(\mathcal{M}^l),
\end{equation} 
where $\mathcal{T} = \{T_i|_{i=0}^{N-1}\}$ are the primary tokens for the video, and $\bigoplus$ denotes the concatenation operation on features from different levels of blocks.
The multi-level feature extraction assures that both low-level features that contains distortion-related information and high-level semantics which is useful for Temporal Content Transformer in Sec. \ref{sec:32} are both captured by $\mathcal{T}$.   




\subsubsection{Temporal Differences} As discussed above, all these temporal distortions come from temporal variations. Though we introduce action recognition backbones to sense these variations, some other distortion-related variations such as unsmooth scene transitions are less related to video actions and are less captured by the backbone. Inspired by \cite{deepvqa,tlvqm}, we extract the temporal differences between features of adjacent frames to further catch the temporal variations and then concatenate them with primary tokens $\mathcal{T}$ to get the STDE tokens $\mathcal{\hat{T}} = \{\hat{T}_i|_{i=0}^{N-1}\}$, defined as follows:
\begin{align} 
    \hat{T}_i &= T_i \oplus (T_i - T_{i+1}),~~~~&0\leq i<N-1 \\
    \hat{T}_{i} &= T_i \oplus \mathbf{0},~~~~&i=N-1
\end{align} where $\oplus$ denotes that two feature vectors are concatenated one after another in the channel dimension.

\subsubsection{Temporal-Distortion-Aware Qualities for Frames} After the token extraction, we build a direct path with a two-layer multi-layer perceptron (MLP) to reduce STDE tokens $\{\hat T_i|_{i=0}^{N-1}\}$ into temporal-distortion-aware qualities for frames ($\{d_i|_{i=0}^{N-1}\}$). Denoting the two linear layers of the MLP as $l_1$ and $l_2$, and the GELU activation function as $\mathrm{A}_\text{GELU}$, the $d_i$ is generated through the following equation:

\begin{equation}
    d_i = l_2(\mathrm{A}_\text{GELU}(l_1(T_i)))
\end{equation}

The $\{\hat T_i|_{i=0}^{N-1}\}$ will also be fed into the TCT to get the temporal quality attention weights and combine with $\{d_i|_{i=0}^{N-1}\}$ to get the final video quality (as in \cref{eq:8}. The structure and pipeline of the TCT is explained in the next section.

We visualize the temporal-distortion-aware frame qualities $\{d_i|_{i=0}^{N-1}\}$ in \cref{sec:vis}.


\subsection {Temporal Content Transformer} \label{sec:32}


We design the temporal content transformer (TCT) to model the content-aware temporal quality attentions of frames with transformers. The TCT first conducts temporal sampling on features (TSF) from the STDE-extracted tokens $\{\hat T_i |_{i=0}^{N-1} \}$ and then reduces the channels of these features. These processed features are passed into a transformer encoder for correlation modeling between different frames, and then a transformer decoder to further learn the correlation of frame content to the overall theme. Finally, an MLP regresses the outputs of the transformers to get the temporal quality attention weights of frames. We discuss each part of the TCT as follows.

\begin{figure}
    \centering
    \includegraphics[width=\linewidth]{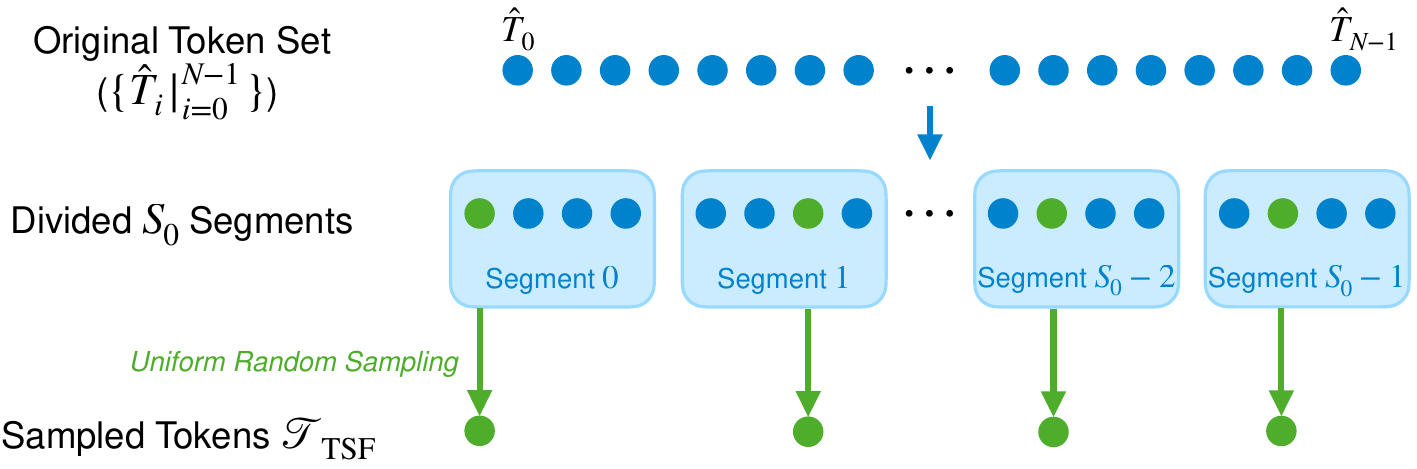}
    \caption{The paradigm of proposed temporal sampling on features (TSF), where the features are divided into $S_0$ uniform segments. We sample multiple times for TSF during inference to improve its stability.}
    \label{fig:tss}
\end{figure}

\subsubsection{Temporal Sampling on Features (TSF)} As transformers need matrix multiplications which is with $O(N^2)$ complexity to sequence length $N$, we use proposed temporally-sampled features instead of the original full features to reduce the input sequence length for the TCT. 
As illustrated in \cref{fig:tss}, we first divide the video into $S_0$ uniform-length segments regardless of the its original length and randomly sample one token from each segment. Given the input $\mathcal{T} = \{\hat T_i |_{i=0}^{N-1}\}$ (tokens extracted from STDE), the token set in temporal sampling on features $\mathcal{T}_\mathrm{TSF}$ is expressed as:  
\begin{equation} \mathcal{T}_\mathrm{TSF} = \{{\hat{T}}_{\mathcal{U}_\text{int}(\frac{j \times N}{S_0}, \frac{(j+1) \times N} {S_0})}|_{j=0}^{S_0 - 1} \}, \end{equation}
where $\mathcal{U}_\text{int}(a, b)$ (where $a < b$) uniformly samples an integer within the range bounded by $a$ and $b$. 

Contrary to the common concern that such sampling will reduce the performance, we notice that for VQA this practice improves the performance in some cases. This is because unlike natural language that is highly abstract, videos are usually continuous between frames and adjacent frames are usually with similar contents. Therefore, the TSF won't lose too much information about the video contents. On the contrary, compared with the vanilla transformer that uses all feature tokens as input, the TSF significantly reduces the token length and thus reduce the training difficulty of the transformer model. Benefited from its randomness, it also increases the total number of training pairs. This is especially helpful and improves the training effectiveness on small datasets such as CVD2014\cite{cvd} and LIVE-Qualcomm\cite{qualcomm} where the TCT tend to overfit, or YouTube-UGC\cite{ytugc} dataset where the inputs are too long. (In \cref{table:tsf1}) The TSF also significantly boosts the training speed and slightly improves the performance on other datasets (In \cref{table:tsf2}). Compared with temporal pooling on inputs, it also keeps the original tokens free from any pooling kernels so that the quality information of frames is not corrupted during the processing. \footnote{During inference, we randomly sample the tokens for $S_m$ multiple times and get the average result of different samples together to improve the prediction stability. Details and analyses can be found in \cref{sec:rely}.}

\subsubsection{Residual Transformer Encoder} 
We introduce a transformer encoder on the TSF-sampled features to learn the content correlations of frames across the whole video. The core module in the transformer encoder is the self-attention module\cite{allyouneed}. Generally, given a sequence of tokens $\mathbf{T} \in \mathbb{R}^{N \times C}$\footnote{where $C$ is the channel number and $N$ is the number of input tokens.} as input, the self-attention module will first project $\mathbf{T}$ into \textit{key, query, value} matrices ($K,Q,V \in \mathbb{R}^{N \times C}$) with matrix multiplications by weights $P_K$, $P_Q$ and $P_V$, as follows:
\begin{equation}
    K = \mathbf{T}P_K, Q = \mathbf{T}P_Q, V = \mathbf{T}P_V
    \label{eq:t1}
\end{equation}
Then, $Q$ will multiply with the transposed $K^T \in \mathbb{R}^{C \times N}$ to get the $M \in \mathbb{R}^{N \times N}$ as follows:
\begin{equation}
M = \text{Softmax}(\frac{QK^T}{\sqrt{C}})
\label{eq:t2}
\end{equation}
and $M^{i,j}$ is the attention value between element $i$ and $j$ in the sequence, reflecting the correlation between them. The computation of the attention value is agnostic to the distance between them and thus especially suitable to model global dependencies for temporal quality attention modeling in VQA.

The proposed transformer encoder ${\Phi}$ contains four sequential layers. Each layer includes a self-attention module as discussed in \cref{eq:t1} and \cref{eq:t2} and several feed-forward MLP layers, following the original structure as proposed in \cite{allyouneed}. We also add a long-range residual link across the transformer encoder to enhance its learning effectiveness, and the whole residual transformer encoder that gets encoded tokens $\mathbf{T}_\text{en} \in\mathbb{R}^{ N\times C}$ from the pre-encoding tokens $\mathbf{T}_\text{pe} \in\mathbb{R}^{ N\times C}$  is expressed as follows:




\begin{equation}
\mathbf{T}_\text{en} = {\Phi}(\mathbf{T}_\text{pe}) + \mathbf{T}_\text{pe} \label{eq:4}
\end{equation} 

\subsubsection{Transformer Decoder} We carefully design a two-layer transformer decoder to detect specific frame contents that catch most human attention in the video, by referring the correlation of a specific frame's content with the average content. As this attention mechanism is related to the correlation of frame content to the overall topic (or theme) of the video, we design to explicitly capture these contents via the cross \textit{key-query} mapping between the encoded token $ \mathbf{T}_\text{en}$ and the average content token $\mathbf{T}_\text{avg}$.  This cross \textit{key-query} mapping is slight different from \cref{eq:t1} and \cref{eq:t2} and formulated as follows.


As the \textit{Decoder Layer} in \cref{fig:DisCoVQA} shows, we perform \textit{temporal pooling} ($\mathcal{P}_t$) on pre-encoding token to get the average token
\begin{equation}
    \mathbf{T}_\text{avg} = \mathcal{P}_t(\mathbf{T}_\text{pe}) \in \mathbb{R}^{ 1\times C}
\end{equation} as the representation of the overall content. $\mathbf{T}_\text{avg}$ is projected into the \textit{query} matrix. The \textit{key} and \textit{value} matrices are both projected from the encoded tokens $ \mathbf{T}_\text{en} \in \mathbb{R}^{ N\times C}$. Then this cross \textit{key-query} mapping gives the $M_\text{QK}$ matrix computed as \cref{eq:crosskeyquerypair}.
\begin{align}
K_\text{de} &= \mathbf{T}_\text{en}P_K, Q_\text{de} = \mathbf{T}_\text{avg}P_Q, V_\text{de} = \mathbf{T}_\text{en}P_V \\
M_\text{QK} &= \text{Softmax}(\frac{Q_\text{de}K_\text{de}^T}{\sqrt{C}})
\label{eq:crosskeyquerypair}
\end{align}


The attention weights $M_\text{QK}$ in the last layer are directly multiplied with the \textit{value} matrix projected from $\mathbf{T}_\text{en}$ and added with the $\mathbf{T}_\text{en}$ with a residual link (similar as in \cref{eq:4}) to get the output token $\mathbf{T}_\text{out}$ of the transformer decoder, which is further reduced to quality attention weights of frames $\{w_i|_{i=0}^{N-1}\}$ with the following scheme (denote the linear layers as $l_3,l_4$).

\begin{equation}
    w = l_4(\mathrm{A}_\text{GELU}(l_3(T_\text{en} + M_\text{QK}V_\text{de})))
\end{equation}

We visualize the learnt $\{w_i|_{i=0}^{N-1}\}$ and $M_\text{QK}$ in \cref{sec:vis}.

\subsection{Final Video Quality Prediction}
\label{sec:33}

The attention weights $w_i$ are multiplied with the disortion-based qualities $d_i$ frame-by-frame and get the final video quality $q$ as follows:

\begin{equation}
    q = \sum_{i=0}^{N-1} \frac{d_i \times (1 + w_i)}{N}
\label{eq:8}
\end{equation}

Following VQEG's suggestions and practices of several existing VQA works~\cite{bvqa2022,vsfa}, we remap the $q$ with consideration of subjective mean opinion scores (MOS) $s$ as follows:
\begin{equation}
\hat q = \frac{\max(s) - \min(s)}{1 + e^{\frac{q - \overline{q}}{\sigma(q)}}} + \min(s)
\end{equation} 
so that the predicted quality scores can be converted to the same range with the corresponding subjective scores. After remapping, we use the MAE loss between $\hat q$ and $s$ 
\begin{equation}
    L = \Vert \hat q - s \Vert_1
\end{equation}
as our training loss function.

\section{Benchmark Experiments}
\label{sec:4}

In this section, we benchmark the performance of DisCoVQA on several natural VQA datasets. The proposed DisCoVQA shows the best performance in most individual dataset evaluation (\cref{sec:intra}), and far better generalization ability in cross-dataset experiments (\cref{sec:inter}). DisCoVQA trained with large VQA datasets LSVQ and KoNViD-150k not only outperforms previous approaches but also shows competitive performance on small benchmark datasets without any fine-tuning process (\cref{sec:large}).  DisCoVQA is also with similar speed to non-transformer baselines (\cref{sec:time}).

\begin{table}[]
    \centering
    \setlength\tabcolsep{2.5pt}
    \caption{Sizes and characteristics of common VQA datasets.}
    \resizebox{\linewidth}{!}{\begin{tabular}{l|c|c|c} \hline
         Dataset & Type & Distortion Type & Size  \\ \hline
         CVD2014 \cite{cvd} & Normal Scale VQA dataset & Simulated Natural & {234} \\
         LIVE-Qualcomm \cite{qualcomm} & Normal Scale VQA dataset & Simluated Natural & {208} \\
         KoNViD-1k \cite{kv1k} & Normal Scale VQA dataset & In-the-wild & {1,200} \\
         LIVE-VQC \cite{vqc} & Normal Scale VQA dataset & In-the-wild & {585} \\ \hdashline
         LSVQ \cite{pvq} & Large Scale VQA dataset & In-the-wild & \red{39,076} \\
         KoNViD-150k \cite{mlsp} &  Large Scale VQA dataset & In-the-wild & \red{150,000} \\ \hline
    \end{tabular}}
    \label{tab:videodatasize}
\end{table}



\begin{table*}[]
\small
\renewcommand\arraystretch{1.35}
\setlength\tabcolsep{2.4pt}
\centering
\caption{Result on CVD2014\cite{cvd}, LIVE-Qualcomm\cite{qualcomm}, KoNViD-1k\cite{kv1k} and LIVE-VQC\cite{vqc} datasets with standard 6:2:2 split setting. Some methods with original results not in this setting are reproduced by us. \textit{Weighted Average} shows the weighted averaged result with respect to dataset sizes and the \textbfred{red} and $\textrm{\blue{blue}}$ results represent the first and second best in chart. Methods with extra-dataset pre-training are labeled with (\textit{EX}), and results that are neither reported nor reproduced are labeled as \gray{NA}.} \label{table:cvdqualcommvqckovid}
\resizebox{\textwidth}{!}{\begin{tabular}{l|ccc|ccc|ccc}
\hline
\textbf{Simulated Natural Datasets} & \multicolumn{3}{c|}{\textbf{CVD2014} (\textit{234}, \cite{cvd})}    & \multicolumn{3}{c|}{\textbf{LIVE-Qualcomm} (\textit{208}, \cite{qualcomm})} &  \multicolumn{3}{c}{\textit{Weighted Average (442)}} \\ \cline{1-4} \cline{5-7} \cline{8-10} 
Method & SROCC(std)                 & PLCC(std)                 & KROCC(std)         & SROCC(std)                 & PLCC(std)                  & KROCC(std)          & SROCC & PLCC & KROCC     \\ \hline
V-BLIINDS\cite{vbliinds}(2012,TIP)        & 0.746(0.056)          & 0.753(0.053) & 0.562(0.057)  & 0.710(0.031)  & 0.704(0.030)          & 0.519(0.026)   & 0.705 & 0.709 & 0.515     \\ 
TLVQM\cite{tlvqm}(2019,TIP)           & 0.830(0.040)          & 0.850(0.040)          & \gray{NA}            & 0.780(0.070)        & 0.810(0.060)          & \gray{NA}  & 0.806 &  0.831 & \gray{NA}    \\ \hdashline
VSFA\cite{vsfa}(2019,MM)            & \blue{0.870(0.037)}         & 0.868(0.032)         & \blue{0.695(0.047)}   & 0.773(0.061)          & 0.795(0.055)          & 0.587(0.062)   & {0.824} & {0.834} & {0.644}    \\ 
3D-CNN+LSTM\cite{cnn+lstm}(2019,ICIP)          & \gray{NA}       & \gray{NA}        & \gray{NA}   & 0.687         & 0.792       & \gray{NA}   & \gray{NA} & \gray{NA} & \gray{NA}   \\
MLSP-FF\cite{mlsp}(2021,Access)        & 0.770(0.060)          & \gray{NA}                   & \gray{NA}            & 0.710(0.080)          & \gray{NA}                   & \gray{NA}   & 0.742 & \gray{NA} & \gray{NA}    \\ 
CNN-TLVQM\cite{cnntlvqm}(2020,MM) & 0.863(0.037) & \blue{0.880(0.025)} & 0.677(0.035) & 0.810(0.045) & \textbfred{0.833(0.029)} & 0.629(0.042) & \blue{0.838} & \blue{0.858} & 0.654  \\
GST-VQA\cite{gstvqa}(2021,TCSVT) & 0.831(0.052) & 0.844(0.062) & 0.657(0.037) & 0.801(0.053) & \blue{0.825(0.043)} & 0.620(0.052) & 0.817 & 0.835 & 0.639 \\
\textit{R+S}(\textit{EX})\cite{bvqa2022}(2022,TCSVT)         &0.860(0.037)&0.877(0.034)&0.687(0.048)&\blue{0.814(0.045)}&0.819(0.054)&\blue{0.639(0.057)}&\blue{0.838}&0.849&\blue{0.664} \\\hline
\textbf{DisCoVQA (Ours)}    & \textbfred{0.897(0.025)} & \textbfred{0.893(0.025)} & \textbfred{0.726(0.033)} & \textbfred{0.823(0.033)} & \blue{0.825(0.030)} & \textbfred{0.645(0.033)} & \textbfred{0.862} & \textbfred{0.862} & \textbfred{0.688} \\ \hline 
\hline
\textbf{Real-world Natural Datasets} & \multicolumn{3}{c|}{\textbf{LIVE-VQC} (\textit{585}, \cite{vqc})}    & \multicolumn{3}{c|}{\textbf{KoNViD-1k} (\textit{1200}, \cite{kv1k})} &  \multicolumn{3}{c}{\textit{Weighted Average (1785)}} \\ \cline{1-4} \cline{5-7} \cline{8-10} 
\text{Method} & SROCC(std)               & PLCC(std)                 & KROCC(std)         & SROCC(std)                & PLCC(std)               & KROCC(std)        & SROCC  & PLCC & KROCC     \\ \hline
V-BLIINDS\cite{vbliinds}(2012,TIP)        & 0.694(0.050)          & 0.718(0.050)          & 0.508(0.042)   & 0.710(0.031)          & 0.704(0.030)          & 0.519(0.026)   & 0.705 & 0.709 & 0.515     \\ 
TLVQM\cite{tlvqm}(2019,TIP)           & 0.799(0.036)          & 0.803(0.036)          & {0.608(0.037)}   & 0.773(0.024)          & 0.768(0.023)          & 0.577(0.022)   & 0.782 & 0.779 & 0.587    \\ 
VIDEVAL\cite{videval}(2021,TIP)        & 0.752(0.039)          & 0.751(0.042)          & 0.564(0.036)   & 0.783(0.021)          & 0.780(0.021)          & 0.585(0.021)   & 0.773 & 0.770 & 0.578   \\ \hdashline
VSFA\cite{vsfa}(2019)(2019,MM)             & 0.773(0.027)          & 0.795(0.026)          & 0.581(0.031)   & 0.773(0.019)          & 0.775(0.019)          & 0.578(0.019)   & 0.773 & 0.782 & 0.579    \\ 

CNN-TLVQM\cite{cnntlvqm}(2020,MM) & 0.814(0.027) & 0.821(0.025) & 0.622(0.033) & 0.816(0.018) & 0.818(0.019) & 0.626(0.018) & 0.815 & 0.819 & 0.625 \\
RAPIQUE\cite{rapique}(2021,OJSP)          & 0.755       & 0.786        & \gray{NA}   & 0.803         & 0.817       & \gray{NA}   & \gray{NA} & \gray{NA} & \gray{NA}   \\
MLSP-FF\cite{mlsp}(2021,Access)         & 0.720(0.060)          & \gray{NA}                   & \gray{NA}            & 0.820(0.020)          & \gray{NA}                   & \gray{NA}            & {0.787} & \gray{NA} & \gray{NA}    \\
CoINVQ\textit{(EX)}\cite{rfugc}(2021,CVPR)          & \gray{NA}       & \gray{NA}        & \gray{NA}   & 0.767         & 0.764       & \gray{NA}   & \gray{NA} & \gray{NA} & \gray{NA}   \\ 
PVQ\textit{(EX)}\cite{pvq}(2021,CVPR)          & {0.803(0.029)}     & {0.811(0.028) }         & {0.616(0.031)}  & 0.785(0.021)         & 0.774(0.028)       & 0.576(0.020)   & 0.791 & 0.786 & 0.589  \\ 
LSCT-PHIQ(\textit{EX})\cite{lsctphiq}(2021,MM)         & 0.796(0.025)         & 0.782(0.024)                 & 0.589(0.023)           & \blue{0.833(0.027)}          & \blue{0.834(0.024)}                   & \blue{0.638(0.019)}             & 0.821 & 0.817 & 0.625    \\
GST-VQA\cite{gstvqa}(2021,TCSVT) & 0.788(0.032) & 0.796(0.028) & 0.604(0.037) & 0.814(0.026) & 0.825(0.043) & 0.621(0.027) & 0.805 & 0.816 & 0.615 \\
\textit{R+S}(\textit{EX})\cite{bvqa2022}(2022,TCSVT)         &\textbfred{0.836(0.031)}&\textbfred{0.831(0.025)}&\textbfred{0.641(0.032)}&0.832(0.023)&0.833(0.019)&0.634(0.017)&\blue{0.833}&\blue{0.832}&\blue{0.636}    \\\hline
\textbf{DisCoVQA (Ours)}    & \blue{0.820(0.030)} & \blue{0.826(0.024)} & \blue{0.633(0.034)} & \textbfred{0.847(0.014)} & \textbfred{0.847(0.028)} & \textbfred{0.660(0.018)} & \textbfred{0.838} & \textbfred{0.840} & \textbfred{0.651} \\ \hline
\end{tabular}}
\end{table*}

%

\subsection{Experimental Settings}
\subsubsection{Implementation Details}
We use one NVIDIA Tesla V100 GPU and Pytorch \cite{pytorch} for training. We set batch size $B=16$, learning rate $lr=0.001$ and use the AdamW \cite{adamw} optimizer with $0.001$ weight decay rate during training. For multi-level feature extraction, we pass original-resolution videos without rescaling them.
Following existing works \cite{vsfa, mdtvsfa, videval, vbliinds}, we use SROCC (Spearman Rank-order Correlation Coefficients), PLCC (Pearson Linear Correlation Coefficients), KROCC (Kendall Rank-order Correlation Coefficients) as our evaluation metrics.
Among them, SROCC \& KROCC reflect the rank correlation of two sequences, where PLCC reflects the linear correlation, and higher correlations means better performances.

\subsubsection{Datasets}
We use six in-the-wild natural VQA datasets, where the majority of videos are directly photographed, instead of generated by algorithms, to benchmark the performance of our model (as listed in \cref{tab:videodatasize}). We select the first four, CVD2014, LIVE-Qualcomm, LIVE-VQC and KoNViD-1k with normal scale to benchmark the effectiveness of our model. These four datasets are further divided into two groups: the datasets with \textbf{simulated distortions} that are common in natural photography, CVD2014 and LIVE-Qualcomm; and the datasets \textbf{collected from real-world} natural videos, LIVE-VQC and KoNViD-1k. We report the \textit{weighted average} performance of each group with respect to their dataset sizes to avoid the random biases during single dataset collections. For the rest two large scale VQA methods proposed recently, we also compare our model with existing approaches on them and evaluate whether directly training on these models can generalize well on other datasets.


\subsubsection{Baseline Methods}
We choose several most recent state-of-the-art methods (and label their time of publication) as comparison. Specifically, we compare with methods that represent different existing temporal modeling strategies in VQA, including VSFA\cite{vsfa}, which applied a ResNet-50 2D-CNN backbone and an RNN for temporal modeling (and GST-VQA\cite{gstvqa} which is based on VSFA and improves the training strategy for VQA); we also compare with CNN-TLVQM\cite{cnntlvqm}, which carefully designed handcrafted features for temporal modeling. We also notice a newly proposed method, MLSP-FF\cite{mlsp} with a heavy CNN backbone and only used naive average pooling for temporal modeling. A very recent approach, \textit{R+S} applies SlowFast\cite{slowfast}, a two-branch 3D-CNN network and a GRU\cite{gru} temporal regression module for temporal modeling, and ensembles it with another spatial branch for VQA.

\subsubsection{Evaluation Settings} \label{sec:intra}
We conduct experiments on CVD2014 \cite{cvd}, LIVE-Qualcomm \cite{qualcomm}, LIVE-VQC \cite{vqc} and KoNViD-1k \cite{kv1k}, four individual VQA benchmark datasets discussed above. We follow the standard 6:2:2 train-validate-test dataset split settings (60\% for training, 20\% for validation, and we report our performance on \textit{the rest 20\% test set} while the validation performance reaches peak), and report the average results on ten random splits for each dataset, together with the standard deviations. This evaluation setting is to avoid overfitted results and improve the confidence of our experimental setting.

\begin{table}[]
\caption{SROCC comparison of different methods with 8:2 setting and 6:2:2 setting on LIVE-VQC and KoNViD-1k.} \label{table:82s}
\setlength\tabcolsep{12pt}
\renewcommand\arraystretch{1.15}
\footnotesize
\center
\resizebox{\linewidth}{!}{\begin{tabular}{l|cc|cc}
\hline
\textbf{Dataset} /         & \multicolumn{2}{|c|}{\textbf{LIVE-VQC}}       &  \multicolumn{2}{|c}{\textbf{KoNViD-1k}}           \\ \cline{2-5}
Method / Split Setting                    & 6:2:2 & 8:2           & 6:2:2 & 8:2                                              \\ \hline 
CNN-TLVQM(\textit{EX})\cite{cnntlvqm}   & 0.830 &  0.814            & 0.830 & 0.816                            \\ \hdashline
PVQ(\textit{EX})\cite{pvq}    &  0.827   & 0.803         & 0.791 & 0.795  \\      \hdashline
LSCT-PHIQ(\textit{EX})\cite{lsctphiq}   & \gray{NA} & 0.796          & 0.860 & 0.833  \\      \hdashline

\textbf{DisCoVQA (Ours)} &  \textbfred{0.838} & \textbfred{0.820} & \textbfred{0.863} & \textbfred{0.847} \\\hline
\end{tabular}}
\end{table}

\begin{table}[]
\caption{PLCC comparison of different methods with 8:2 setting and 6:2:2 setting on LIVE-VQC and KoNViD-1k.} \label{table:82p}
\setlength\tabcolsep{12pt}
\renewcommand\arraystretch{1.15}
\footnotesize
\center
\resizebox{\linewidth}{!}{\begin{tabular}{l|cc|cc}
\hline
\textbf{Dataset} /         & \multicolumn{2}{|c|}{\textbf{LIVE-VQC}}       &  \multicolumn{2}{|c}{\textbf{KoNViD-1k}}           \\ \cline{2-5}
Method / Split Setting                    & 6:2:2 & 8:2           & 6:2:2 & 8:2                                              \\ \hline 
CNN-TLVQM(\textit{EX})\cite{cnntlvqm}   & 0.840 &  0.821            & 0.830 & 0.818                            \\ \hdashline
PVQ(\textit{EX})\cite{pvq}    &  0.837   & 0.811         & 0.786 & 0.774  \\      \hdashline
LSCT-PHIQ(\textit{EX})\cite{lsctphiq}   & \gray{NA} & 0.782          & 0.850 & 0.834  \\      \hdashline

\textbf{DisCoVQA (Ours)} &  \textbfred{0.844} & \textbfred{0.826} & \textbfred{0.860} & \textbfred{0.847} \\\hline
\end{tabular}}
\end{table}

\begin{table*}[]
\setlength\tabcolsep{5.2pt}
\renewcommand\arraystretch{1.1}
\caption{Cross-dataset results: between LIVE-VQC \cite{vqc}, KoNViD-1k \cite{kv1k} and YouTube-UGC \cite{ytugc}. Results of DisCoVQA related to YouTube-UGC \cite{ytugc} is conducted with the 236-fewer-video version. We still reach better cross-dataset performance when training on YouTube-UGC with less videos (our results related to YouTube-UGC are labeled with $^\star$ and presented only for reference due to video missing.)} \label{table:cross}
\center
\footnotesize
\resizebox{\linewidth}{!}{\begin{tabular}{l|cc|cc|cc|cc|cc|cc}
\hline
\textbf{Train} on & \multicolumn{4}{c|}{\textbf{KoNViD-1k}} & \multicolumn{4}{c|}{\textbf{LIVE-VQC}} & \multicolumn{4}{c}{\textbf{Youtube-UGC}} \\ \hline
\textbf{Test} on & \multicolumn{2}{c|}{\textbf{LIVE-VQC}} & \multicolumn{2}{c|}{\textbf{Youtube-UGC}}  & \multicolumn{2}{c|}{\textbf{KoNViD-1k}} & \multicolumn{2}{c|}{\textbf{Youtube-UGC}}   & \multicolumn{2}{c|}{\textbf{LIVE-VQC}} & \multicolumn{2}{c}{\textbf{KoNViD-1k}} \\ \hline
\textit{}                         & SROCC                        & PLCC                        & SROCC                        & PLCC                 & SROCC                        & PLCC                        & SROCC                        & PLCC        & SROCC                        & PLCC               & SROCC                        & PLCC                                          \\ 
\hline
CNN-TLVQM\cite{cnntlvqm}             & 0.713                      & 0.752    & \gray{NA} &   \gray{NA}               & 0.642 & 0.631 & \gray{NA} & \gray{NA}       & \gray{NA}                        & \gray{NA}             & \gray{NA}  &   \gray{NA}                                 \\ 
GST-VQA\cite{gstvqa}             & 0.700                      & 0.733    & \gray{NA} &   \gray{NA}               & \blue{0.709} & 0.707 & \gray{NA} & \gray{NA}       & \gray{NA}                        & \gray{NA}             & \gray{NA}  &   \gray{NA}                                 \\ 
VIDEVAL\cite{videval}                    & 0.627                         & 0.654   & 0.370 & 0.390                  & 0.625                        & 0.621 & 0.302 & 0.318  & 0.542                         & 0.553      & 0.610 &   0.620                           \\ 
MDTVSFA\cite{mdtvsfa}                     & \blue{0.716}                        & \blue{0.759}    & \blue{0.408} &  \blue{0.443}                 & 0.706                         & \blue{0.711}  & \blue{0.355} &  \blue{0.388}      & \blue{0.582}                        & \blue{0.603}   & 0.649 &   0.646                          \\ \hline
\textbf{DisCoVQA (Ours)}             & \textbfred{0.782}               & \textbfred{0.797}       & \textbfred{0.415$^\star$} & \textbfred{0.449$^\star$}    & \textbfred{0.792}               & \textbfred{0.785}  & \textbfred{0.409$^\star$} & \textbfred{0.432$^\star$}   & \textbfred{0.661$^\star$}               & \textbfred{0.685$^\star$}       &  \textbfred{0.686$^\star$} &   \textbfred{0.697$^\star$}                 \\ \hline
\end{tabular}}
\end{table*}

\subsection{Comparison on Individual Datasets}

As \cref{table:cvdqualcommvqckovid} shows, our proposed DisCoVQA has shown superior performance than existing methods published prior to us on four individual benchmark datasets, even those with extra-dataset pretraining (labeled as \textit{EX}). For example, the proposed model is up to \textbf{9.5\%} better than VSFA\cite{vsfa} with similar parameters and computational complexities. The proposed model is also up to \textbf{13.8\%} better than MLSP-FF which has much more parameters but no temporal modeling other than average pooling, showing the vitality of proper temporal modeling in VQA.

The proposed model is \textbf{state-of-the-art on three datasets} and the runner-up model on LIVE-VQC, slightly inferior to \textit{R+S}\cite{bvqa2022}. However, \textit{R+S} includes an additional spatial branch that is fine-tuned on other IQA datasets. The proposed model is \textbf{14.5\%}\footnote{Result based on their paper. Same for the next.} better than its pure-temporal branch and \textbf{5.7\%} better than it while their spatial branch is not extra-pretrained. This comparison suggests that at least the proposed transformer-based approach is competitive for temporal modeling. \footnote{Some methods did not provide their codes or report their full performance, so we directly reported results in their paper and left their missing results empty. We try our best to reproduce and fill in every comparison for every competitive method.} 


We also notice different datasets, though all aimed at collecting natural distortions, have different characteristics, especially considered in the temporal domain. For example, the KoNViD-1k contains more \textbf{diverse contents across time} and \textbf{compression-based artifacts}, while LIVE-VQC contains more \textbf{in-capture temporal distortions}. Therefore, the \textit{weighted average} performance on these two datasets (\cref{table:cvdqualcommvqckovid}, in the rightmost) might be a more reliable benchmark to evaluate the full ability of methods. The proposed DisCoVQA reaches state-of-the-art on group weighted averages for both simulated and real-world natural datasets, showing that the proposed method is generally robust rather specially effective on some specific distortion types. 
 
We also notice that some recent deep methods \cite{cnntlvqm, pvq, lsctphiq} directly report the performance on the validation set in a different 8:2 setting (80\% for training, 20\% for validation, no extra testing dataset). To evaluate the difference of our setting and this setting, we also evaluated the proposed DisCoVQA on this setting and try our best to reproduce them in our setting. The results in \cref{table:82p} and \cref{table:82s} suggested that two different settings have statistically prominent performance gap, indicating that we might need to align these settings before making fair comparisons between different methods.

\subsection{Cross-dataset Results} \label{sec:inter}

To measure the generalization ability of the proposed DisCoVQA, we compare cross-dataset results with several state-of-the-art methods.  We carefully choose the methods with relatively good generalization ability for comparison: MDTVSFA \cite{mdtvsfa} is specially designed for multi-dataset alignment; GST-VQA\cite{gstvqa} is also designed for better generalization; CNN-TLVQM \cite{cnntlvqm}, CoINVQ \cite{rfugc} and VIDEVAL \cite{videval} ensemble different types of features for robustness. Without including any special design, the proposed DisCoVQA reaches better generalization than them. The cross-dataset results among KoNViD-1k \cite{kv1k}, LIVE-VQC \cite{vqc} and YouTube-UGC \cite{ytugc}\footnote{The \textit{$^\star$-labeled} results related to YouTube-UGC are shown for reference only due to difference of dataset size (our used are 236 fewer than the original version due to missing of the downloadable links).} are reported in \cref{table:cross}.

Compared with existing methods, we observe more obvious improvements in cross-dataset experiments. Take our comparison with CNN-TLVQM \cite{cnntlvqm} as an example. CNN-TLVQM only relies on handcraft features for temporal modeling, while we design both temporal distortion and temporal attention modeling with transformer-based backbones for it. As \cref{table:cvdqualcommvqckovid} and \cref{table:cross} shows, we outperform CNN-TLVQM by \textbf{10\%} in cross-dataset results (between LIVE-VQC \& KoNViD-1k) where the proposed model is only 2\% better than it during intra-dataset settings. It demonstrates that though applying handcraft or other traditional solutions to tackle the temporal relationships can reach good performance, the proposed transformer-based approach can learn more these relationships more robustly.

We also notice that current methods still cannot generalize well between KoNViD-1k/LIVE-VQC and YouTube-UGC, which might be due to the large proportion of non-photographic videos (categories \textit{games}, \textit{animation}, \textit{lyric videos}, \textit{news report} in it) in YouTube-UGC, while there are very few of them in LIVE-VQC and KoNViD-1k. It will be a nice future objective to improve the generalization ability of VQA approaches between these generated videos and other natural videos.

\subsection{Results on Large-scale Datasets} \label{sec:large}

We evaluate the proposed model on two large-scale datasets, LSVQ \cite{pvq} and KoNViD-150k \cite{mlsp}. As \cref{table:large} shows, we outperform the PVQ \cite{pvq} with \textbf{7.7\%} improvement on the same setting, and 5.1\% improvement even when PVQ uses the additional `\textit{patch}' annotations. The advanced performance of DisCoVQA has shown the effectiveness of the proposed temporal modeling methodology. We also outperform the only available model trained on KoNViD-150k, the MLSP-FF \cite{mlsp} which proposed the dataset and VSFA (reproduced by us).

\begin{figure}[t]
    \centering
    \includegraphics[width=1.0\linewidth]{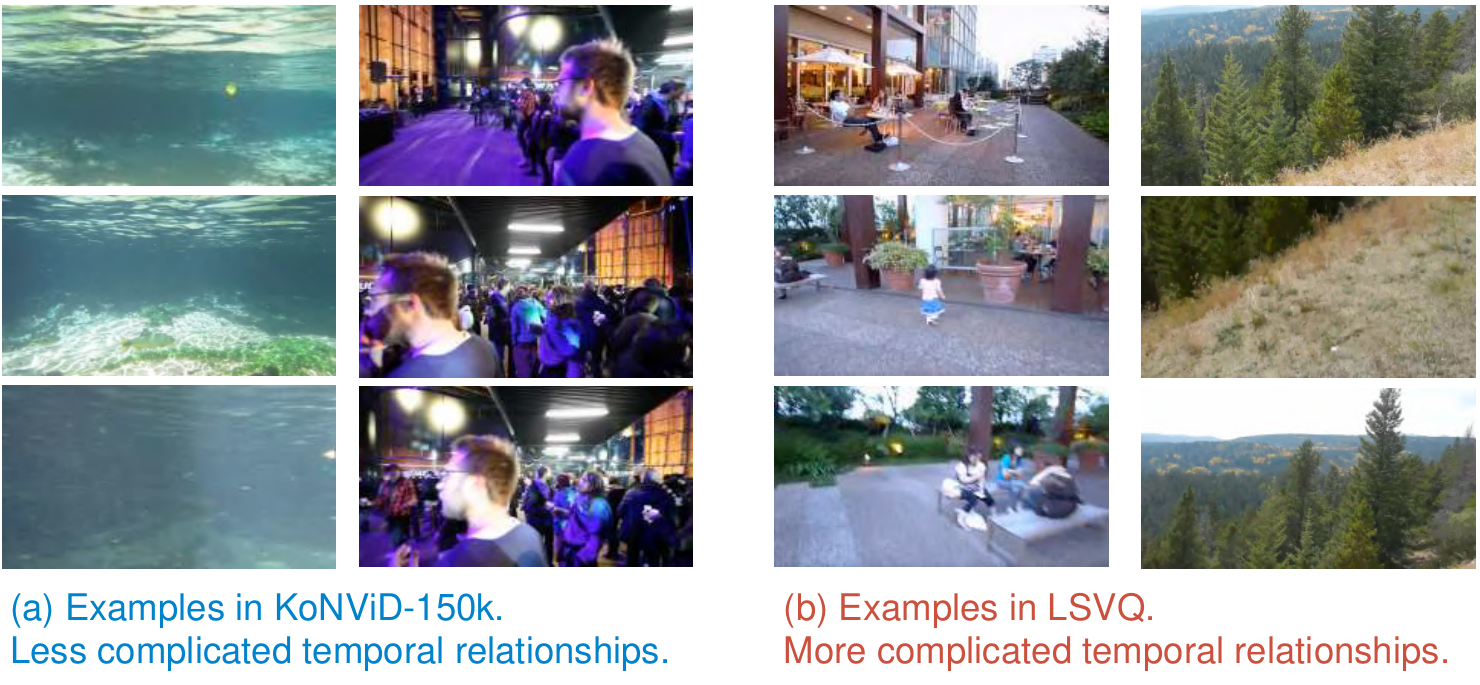}
    \caption{Comparison of (a) KoNViD-150k \cite{mlsp}  and (b) LSVQ \cite{pvq}. The LSVQ dataset has shown more complicated temporal relationships (more difference between frame contents; more temporal distortions) than the KoNViD-150k.}
    \label{fig:largedataset}
\end{figure}

We further notice that the cross-dataset performance on KoNViD-1k \cite{kv1k} and LIVE-VQC \cite{vqc} of DisCoVQA trained with LSVQ is obviously better than that of KoNViD-150k. We suspect that this might be due to the different temporal relationships in two datasets. As illustrated in \cref{fig:largedataset}, the temporal relationships (both temporal distortions and different contents of different frames) are less complicated in KoNViD-150k than in LSVQ, which might be caused by the different collection sources of these two datasets and the different average durations (KoNViD-150k: 5s, LSVQ: \~ 7.5s). The LSVQ dataset with more complicated temporal relationships have reached better generalization ability in natural VQA datasets, suggesting that temporal issues are common in natural videos.

\begin{table}[]
\caption{Large Dataset Training I: pretrained with large-scale LSVQ \cite{pvq} and evaluate on different sets without fine-tuning.} \label{table:large}
\renewcommand\arraystretch{1.2}
\center
\footnotesize
\resizebox{\linewidth}{!}{\setlength\tabcolsep{2.0pt}
\begin{tabular}{l|cc|cc|cc}
\hline
\textbf{Train} on & \multicolumn{6}{c}{\textbf{LSVQ}$_\text{train}$} \\ \hline
\textbf{Test} on & \multicolumn{2}{c|}{\textbf{LIVE-VQC}} & \multicolumn{2}{c|}{\textbf{KoNViD-1k}} & \multicolumn{2}{c}{\textbf{LSVQ}$_\text{test}$} \\ \cline{1-7} 
\textit{~}  & SROCC                        & PLCC                        & SROCC                        & PLCC  & SROCC & PLCC           \\ \hline
VSFA\cite{vsfa}                       & 0.734 & 0.772 & 0.784 & 0.794  & 0.801 & 0.796      \\ 
TLVQM\cite{tlvqm}                       & 0.670 & 0.691 & 0.732  & 0.724  & 0.772 & 0.774       \\ 
PVQ\cite{pvq}                    & 0.747 & 0.776 & 0.781                         & 0.781  & 0.814 & 0.816 \\ 
\textit{PVQ+}                   & \blue{\textit{0.770}} & \blue{\textit{0.807}}       & \blue{\textit{0.791}}       & \blue{\textit{0.795}}   &  \blue{\textit{0.827}}  & \blue{\textit{0.828}} \\ \hline

\textbf{DisCoVQA (Ours)}            &  \textbfred{0.823} & \textbfred{0.837}     &   \textbfred{0.846} & \textbfred{0.849} & \textbfred{0.859} & \textbfred{0.850}   \\ \hline
\end{tabular}}
\end{table}

\begin{table}[]
\caption{Large Dataset Training II: pretrained with large-scale KoNViD-150k \cite{mlsp} and evaluate on different sets without fine-tuning.} \label{table:large2}
\setlength\tabcolsep{2.0pt}
\renewcommand\arraystretch{1.2}
\center
\footnotesize
\resizebox{\linewidth}{!}{\begin{tabular}{l|cc|cc|cc}
\hline
\textbf{Train} on &  \multicolumn{6}{c}{\textbf{KoNViD-150k-A}} \\ \hline
\textbf{Test} on & \multicolumn{2}{c|}{\textbf{LIVE-VQC}} & \multicolumn{2}{c|}{\textbf{KoNViD-1k}} & \multicolumn{2}{c}{\textbf{KoNViD-150k-B}} \\ \cline{1-7} 
\textit{~} & SROCC                        & PLCC                        & SROCC                        & PLCC  & SROCC & PLCC                        \\ \hline
VSFA\cite{vsfa}                       & 0.708 & 0.733 & 0.801 & 0.815  & 0.813 & 0.808    \\
MLSP-FF\cite{mlsp} & \blue{0.738} & \blue{0.754} & \blue{0.828} & \blue{0.821} & \blue{0.827} & \blue{0.852}  \\\hline
\textbf{DisCoVQA (Ours)}   & \textbfred{0.751} & \textbfred{0.766}     &   \textbfred{0.843} & \textbfred{0.841} & \textbfred{0.845} & \textbfred{0.858}        \\ \hline
\end{tabular}}
\end{table}

\begin{table}[t]
\caption{Running time comparison for DisCoVQA on a 540P, 8-second video in KoNViD-1k\cite{kv1k} dataset. The standard deviants for the corresponding running times are in brackets.} \label{table:cost}
\setlength\tabcolsep{3.2pt}
\renewcommand\arraystretch{1.3}
\footnotesize
\center
\resizebox{\linewidth}{!}{\begin{tabular}{l|c|c|c}
\hline
\textbf{Backbone Extractor} /         & ResNet-50 & I3D-ResNet-50  & Swin-T (as proposed)    \\ \cline{2-4} 
Running Time & 2.71s(0.14s) & 2.45s(0.17s) & \textbf{2.33s(0.09s)}  \\
\hline
\textbf{Temporal Regression} /         & 6-layer GRU & vanilla Transformer  & the TCT (proposed)    \\ \cline{2-4} 
Running Time  & 0.053s(0.011s) & 0.120s(0.018s) & \textbf{0.021s(0.004s)} \\
\hline

\end{tabular}}
\end{table}

\subsection{Running Time Comparison}
\label{sec:time}
We discuss the computational cost introduced by transformers in two parts: the running time for the transformer-based backbone (Swin-T) in the STDE and for the transformer-based TCT, and compare with several existing alternative approaches (CNN\&RNN), as shown in \cref{table:cost}. As the results shows, switching the traditional CNN/3D-CNN backbones into the transformer-based Swin-T backbone does not lead to additional running time. The proposed TCT with the temporal sampling on features (TSF) is also \textbf{6x} faster than vanilla transformer and \textbf{2.5x} faster than GRU, proving that it alleviates the problem of high computation loads of transformers on long sequences. These results prove that the proposed DisCoVQA maintains the efficiency of existing deep VQA models while reaching remarkably better performance with transformer-based architectures.


\section{Experimental Analysis on Model Design}
\label{sec:4}

In this part, we would like to answer three important questions to evaluate the effectiveness of the proposed model:

\begin{enumerate}
    \item Does every design of the proposed model lead to reasonable performance improvement? (In \cref{sec:ablstde} and \cref{sec:abltct})
    \item In which particular cases can the model show better ability? (In \cref{sec:vis})
    \item Does the model give reliable results as the TSF includes some randomness during inference? (In \cref{sec:rely}.)
\end{enumerate}

Without special notes, the training datasets for all these studies are LSVQ dataset due to its large scale (28K training videos) and diversity on temporal relationships (as discussed in \cref{sec:large}).

\subsection{Ablation Studies on the STDE} \label{sec:ablstde}

To discuss the effectiveness of the proposed STDE, we run ablation studies on LSVQ dataset \cite{pvq}, and provide result comparisons on both intra-dataset and cross-dataset (test on LIVE-VQC and KoNViD-1k) experiments. We run ablation studies for both the backbone network and micro-designs (multi-level extraction \& temporal differences) in the STDE.

\subsubsection{Effects of Backbone Structures} In \cref{table:backbone}, we replace the Swin-T backbone into two CNN backbones with similar parameters and running time (as compared in \cref{sec:time}): the ResNet-50 that does not extract temporal quality information, and the I3D-ResNet-50 that extracts temporal quality information with convolution kernels. The I3D-ResNet-50 backbone is better than ResNet-50 backbone as it has temporal sensitivity, but ours with Swin-T still performs notably better than it, especially on LIVE-VQC test set with most severe temporal distortions. It should also be noted that even with the same backbone (ResNet50), our method still has better performance than VSFA, proving that our designs other than the backbone are also effective.

\begin{table}[]
\caption{Ablation study on backbone structures in the STDE. All these backbones have similar parameters and running time for feature extraction (\cref{table:cost}).} \label{table:backbone}
\setlength\tabcolsep{2.4pt}
\renewcommand\arraystretch{1.25}
\footnotesize
\center
\resizebox{\linewidth}{!}{\begin{tabular}{l|cc|cc|cc}
\hline
\textbf{Testing} on /         & \multicolumn{2}{|c|}{\textbf{LSVQ$_\text{test}$}}        &  \multicolumn{2}{|c|}{\textbf{KoNViD-1k}}  & \multicolumn{2}{|c}{\textbf{LIVE-VQC}}             \\ \cline{2-7}
Backbone Network                    & SROCC & PLCC           & SROCC & PLCC                      & SROCC & PLCC                               \\ \hline 
\gray{VSFA\cite{vsfa} (with ResNet-50)}        & 0.801 &  0.796             & 0.784  &  0.794                      & 0.734 &    0.772            \\                  \hline 
ResNet-50\cite{he2016residual}             & 0.823  &  0.822           & 0.810 &  0.815                       & 0.758  &  0.776                               \\
  I3D-ResNet-50\cite{k400}             & 0.840  &  0.832           & 0.825 &  0.817                       & 0.774  &  0.793                               \\
\hdashline
\textbf{Swin-T} (as proposed) &  \textbf{0.859} & \textbf{0.850} & \textbf{0.846} & \textbf{0.849}         &  \textbf{0.823} & \textbf{0.837} \\ \hline

\end{tabular}}
\end{table}

\subsubsection{Effects of Micro-designs: Multi-level \& Temporal Differences} We discuss about the two important designs in the STDE: the multi-level feature extraction and the temporal differences. As shown in \cref{table:microdesign}, the multi-level extraction significantly improves the accuracy on all testing scenarios, supporting our claim that it enhances the distortion sensitivity of the STDE. The temporal differences also show non-trivial improvements on all sets, and is especially helpful (+2\%) on LIVE-VQC test set with all hand-held video shots, where the temporal distortions are most severe. The combination of these two designs lead to 3\% improvement than the vanilla Swin-T extractor (\textit{w/o} Both in the Table). Both micro-designs help the proposed transformer-based STDE to be more suitable for extracting temporal distortions.

\begin{table}[]
\caption{Ablation study on multi-level extraction and temporal differences. Both micro-designs further improves the performance of transformer-based STDE.} \label{table:microdesign}
\setlength\tabcolsep{2.4pt}
\renewcommand\arraystretch{1.25}
\footnotesize
\center
\resizebox{\linewidth}{!}{\begin{tabular}{l|cc|cc|cc}
\hline
\textbf{Testing} on /         & \multicolumn{2}{|c|}{\textbf{LSVQ$_\text{test}$}}        &  \multicolumn{2}{|c|}{\textbf{KoNViD-1k}}  & \multicolumn{2}{|c}{\textbf{LIVE-VQC}}             \\ \cline{2-7}
Micro-Design                    & SROCC & PLCC           & SROCC & PLCC                      & SROCC & PLCC                               \\ \hline 

\textit{w/o} Both (remove micro-designs)   & 0.828  &  0.824           & 0.819 &  0.824                       & 0.786  &  0.798                               \\ \hdashline
\textit{w/o} Multi-level Extraction   & 0.835  &  0.829           & 0.825 &  0.828                       & 0.794  &  0.809                               \\

\textit{w/o} Temporal Differences   & 0.844  &  0.835           & 0.836  &  0.839                      & 0.806  & 0.822                                 \\ \hdashline

\textbf{Full STDE} (proposed) &  \textbf{0.859} & \textbf{0.850} & \textbf{0.846} & \textbf{0.849}         &  \textbf{0.823} & \textbf{0.837} \\ \hline

\end{tabular}}
\end{table}

\subsection{Ablation Studies on the TCT} \label{sec:abltct}

In this part, we compare the proposed TCT with two groups of variants on LSVQ dataset. The first group of variants are non-transformer structures: the temporal MLP, the temporal CNN, and the LSTM (a type of RNN); the second group of variants are the structural variants of transformers. We also compare the model variant that removes the TCT at all. Moreover, we discuss the effects for the temporal sampling on features (TSF) for training the TCT on different scale of datasets to show how they help to improve the hard cases of introducing transformers in VQA.

\subsubsection{Comparison with Non-Transformer Structures} In this part, we compare the transformer-based TCT with several non-transformer structures. We set these structures with the same layers (6 layers) as the TCT to make fair comparisons and the results are shown in \cref{table:tctdesign}, Group 1. The proposed transformer-based TCT has much better performance than all non-transformer variants and the variant that removes the TCT, demonstrating that transformer architectures are better for temporal quality attention modeling in VQA.

\subsubsection{Comparison with Transformer-based Variants} In this part, we compare the proposed encoder-decoder-like TCT with several variants, including the pure transformer encoder with 4/6 layers and the variant that changes the average content token ($\mathbf{T}_\text{avg}$) into a zero token as the target input of the transformer decoder. Changing the proposed structure into all these variants result in \textbf{2\%} performance drop, proving the effectiveness of the proposed encoder-decoder-like structure of the TCT which takes the average content as target input. This result also suggests that the overall content is vital in deciding the temporal quality attention across frames in VQA.

\begin{table}[]
\caption{Ablation study of the TCT architecture: compared with non-transformer structures and structural variants of transformers. The corresponding running time comparison can be found in \cref{table:cost}.} \label{table:tctdesign}
\setlength\tabcolsep{1.8pt}
\renewcommand\arraystretch{1.3}
\footnotesize
\center
\resizebox{\linewidth}{!}{\begin{tabular}{l|cc|cc|cc}
\hline
\textbf{Testing} on /         & \multicolumn{2}{|c|}{\textbf{LSVQ$_\text{test}$}}        &  \multicolumn{2}{|c|}{\textbf{KoNViD-1k}}  & \multicolumn{2}{|c}{\textbf{LIVE-VQC}}             \\ \cline{2-7}
Variant                   & SROCC & PLCC           & SROCC & PLCC                      & SROCC & PLCC                               \\ \hline 
\textit{remove the TCT} & 0.842 & 0.836 & 0.826 & 0.823 & 0.804 & 0.813  \\ \hdashline
 \multicolumn{7}{l}{Group 1: Non-Transformer Structures} \\ \hdashline
Temporal MLP (6-layer)     &  0.836 &   0.828          & 0.824 &   0.818                      &  0.792  &  0.801             \\

Temporal CNN (6-layer)    & 0.839 & 0.832 & 0.828 & 0.822 & 0.799 & 0.808 \\
 
LSTM (RNN, 6-layer)    &  0.841 &  0.838    & 0.831  &  0.830   &  0.806 &  0.816                             \\  \hline
 \multicolumn{7}{l}{Group 2: Structural Variants of Transformers} \\ \hdashline
Pure Encoder (4-layer) &  0.847 &   0.840          & 0.837 &   0.838                      &  0.812  &  0.826             \\
Pure Encoder (6-layer) &  0.848 &   0.842          & 0.841 &   0.842                      &  0.812  &  0.828             \\
\textit{change $\mathbf{T}_\text{avg}$ as zero token} & 0.852 &   0.845          & 0.843 &   0.844                      &  0.813  &  0.832  \\ \hline

\textbf{Full TCT} (proposed) &  \textbf{0.859} & \textbf{0.850} & \textbf{0.846} & \textbf{0.849}         &  \textbf{0.823} & \textbf{0.837} \\ \hline

\end{tabular}}
\end{table}

\begin{figure*}
    \centering
    \includegraphics[width=0.88\linewidth]{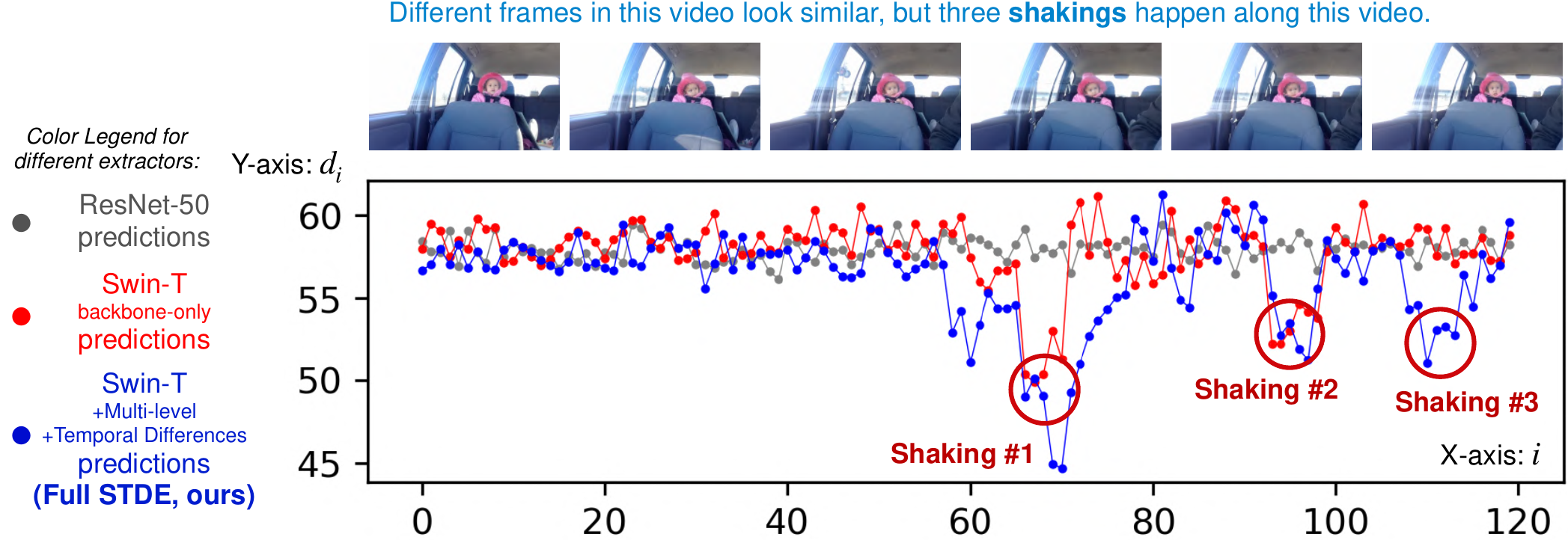}
    \caption{Visualizations of temporal-distortion-aware qualities $d_i$ learnt in STDE, compared with different variants of STDE. Discussions are in \cref{sec:vis}.}
    \label{fig:stdevis}
\end{figure*}

\begin{figure*}
    \centering
    \includegraphics[width=\linewidth]{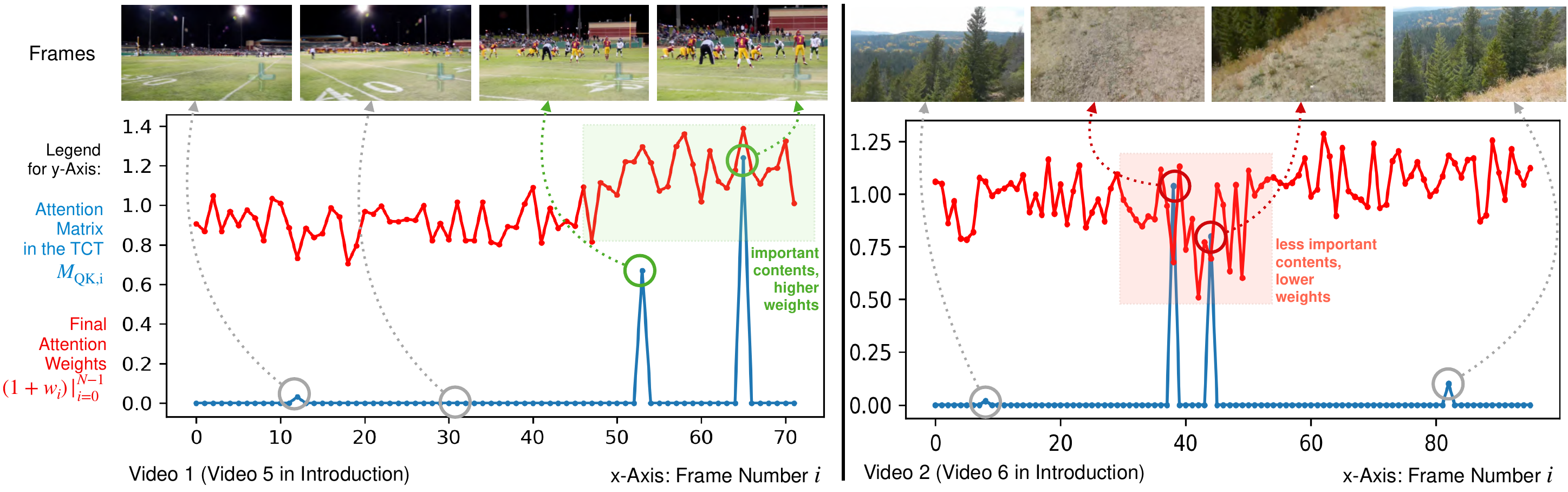}
    \caption{Visualizations of temporal quality attention, including the attention matrix in TCT and the final temporal quality attention weights,
    demonstrating that the propose TCT can learn reasonable temporal quality attention. Discussions are in \cref{sec:vis}.}
    \label{fig:df}
\end{figure*}

\begin{table}[]
\caption{Ablation study of the proposed temporal sampling on features (TSF) on small and long-duration VQA datasets.} \label{table:tsf1}

\setlength\tabcolsep{2.3pt}
\renewcommand\arraystretch{1.25}
\footnotesize
\center
\resizebox{\linewidth}{!}{\begin{tabular}{l|cc|cc|cc}
\hline
\textbf{Dataset} /         & \multicolumn{2}{c|}{\textbf{CVD2014}}        &  \multicolumn{2}{c|}{\textbf{LIVE-Qualcomm}} & \multicolumn{2}{c}{\textbf{YouTube-UGC}}          \\ \cline{2-7}
\textbf{Size/Average Duration}       & \multicolumn{2}{c|}{\textbfred{234}/10s}        &  \multicolumn{2}{c|}{\textbfred{208}/\textbfred{15s}} & \multicolumn{2}{c}{1144/\textbfred{20s}}          \\ \cline{2-7}   
Strategy                    & SROCC & PLCC & SROCC & PLCC   & SROCC    & PLCC                           \\ \hline 
full-length features (\textit{w/o} TSF)      & 0.878 & 0.879  & 0.803 & 0.805 & 0.789  & 0.790                            \\
pooling on features      & 0.883 & 0.880  & 0.799 & 0.801 & 0.794  & 0.797                            \\ \hdashline
sampling on features (proposed)  & \textbf{0.897} & \textbf{0.893} & \textbf{0.823} & \textbf{0.825} & \textbf{0.809} &  \textbf{0.808} \\ \hline

\textit{improvement} to \textit{w/o} TSF & \textbf{+2.2}\% & \textbf{+1.6}\% & \textbf{+2.6}\% & \textbf{+2.9}\% & \textbf{+2.5}\% &  \textbf{+2.3}\% \\ \hline

\end{tabular}}
\end{table}
\begin{table}

\caption{Ablation study of the TSF on other datasets.} \label{table:tsf2}

\setlength\tabcolsep{2.5pt}
\renewcommand\arraystretch{1.25}
\footnotesize
\center
\resizebox{\linewidth}{!}{\begin{tabular}{l|cc|cc|cc}
\hline
\textbf{Dataset} /         & \multicolumn{2}{c|}{\textbf{LIVE-VQC}}        &  \multicolumn{2}{c|}{\textbf{KoNViD-1k}} & \multicolumn{2}{c}{\textbf{LSVQ}}          \\ \cline{2-7}
\textbf{Size/Average Duration}       & \multicolumn{2}{c|}{{585}/8s}        &  \multicolumn{2}{c|}{1200/10s} & \multicolumn{2}{c}{39075/7s}          \\ \cline{2-7}   
Strategy                    & SROCC & PLCC & SROCC & PLCC   & SROCC    & PLCC                           \\ \hline 
full-length features (w/o TSF)      & 0.816 & 0.823  & 0.846 & 0.847 & \textbf{0.860}  & \textbf{0.850}                            \\
pooling on features      & 0.809 & 0.818  & 0.839 & 0.840 & 0.857  & 0.846                          \\ \hdashline
sampling of features (proposed)  & \textbf{0.820} & \textbf{0.826} & \textbf{0.847} & \textbf{0.847} & 0.859 &  \textbf{0.850}\\ \hline

\textit{improvement} to \textit{w/o} TSF & \textbf{+0.4}\% & \textbf{+0.3}\% & \textbf{+0.1}\% & {0.0}\% & {-0.1}\% &  {0.0}\% \\ \hline

\end{tabular}}
\end{table}

\subsubsection{Effects of temporal sampling on features (TSF)}  We show the effectiveness of implementing the TSF for the TCT regression in several different datasets. First, as shown in \cref{table:tsf1}, for datasets that are either small (LIVE-Qualcomm and CVD2014) or with long duration (YouTube-UGC), the TSF significantly helps the learning process of the TCT. It is also noteworthy that on other datasets (as compared in \cref{table:tsf2}), the TSF does not lead to noticeable better performance. As TSF also significantly improves the training speed, we still take it as a part of the TCT when training DisCoVQA in these datasets. Also, the temporal pooling on features consistently perform worse than the TSF on six datasets, proving that keeping original features is important to the learning of TCT. The TSF provides a feasible way of implementing transformers into VQA both effectively and efficiently.

\begin{figure*}
    \centering
    \includegraphics[width=1.0\linewidth]{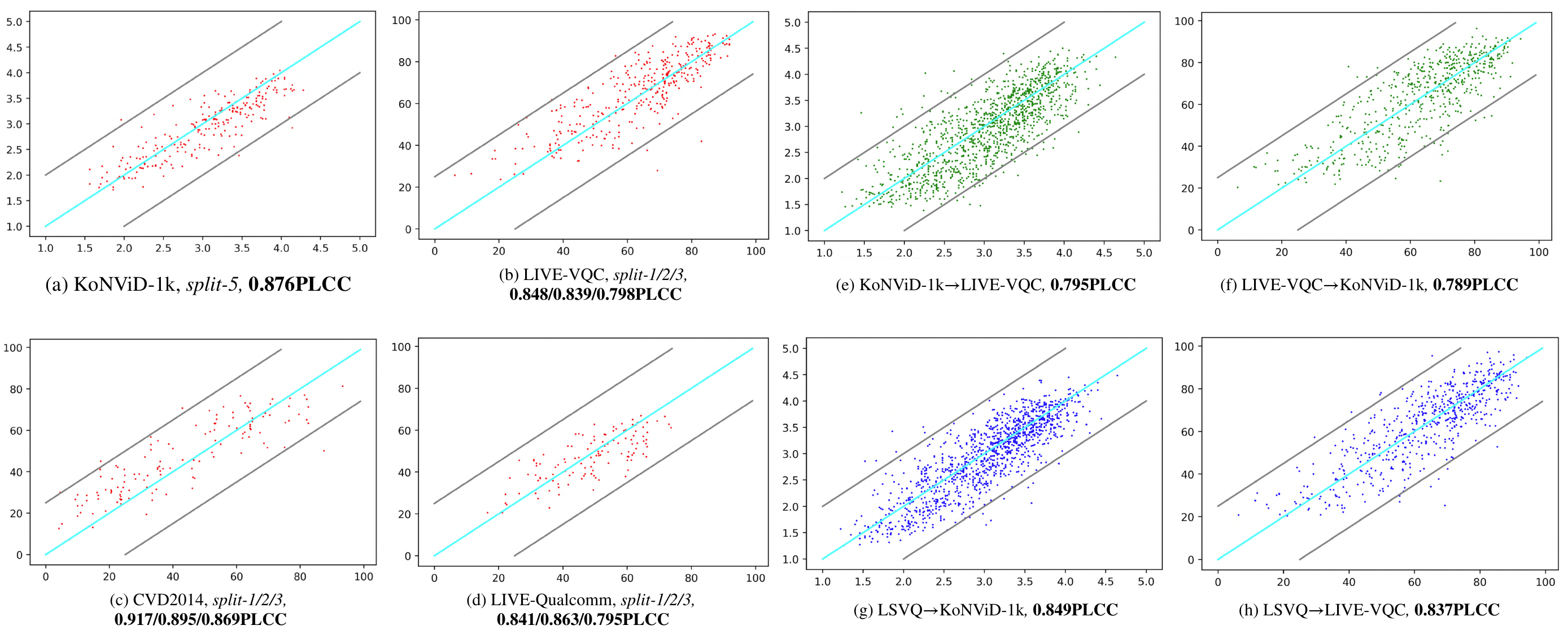}
    \caption{The correlation of predicted scores (\textbf{y-axis}) and the ground truth labels (\textbf{x-axis}). (a)(b)(c)(d) show settings within individual datasets, and (e)(f)(g)(h) show cross-dataset settings. Corresponding benchmarks are in \cref{table:cvdqualcommvqckovid}, \cref{table:cross} and \cref{table:large}.}
    \label{fig:corr}
\end{figure*}

\subsection{Qualitative Results}
\label{sec:vis}

\subsubsection{Visualization of Temporal Distortion Sensitivity} 

For the STDE, we aim to spot the temporal distortions between frames, such as shaking and flickers in videos. In \cref{fig:stdevis}, we visualize the distortion-only qualities of frames in a video that contain shaking learnt with features extracted by different approaches. As illustrated in the figure, compared to vanilla feature extraction from CNN-based backnones, the adoption of Swin-T backbone and two micro-designs in STDE both enhances the temporal distortion extraction by detecting more shakings, and helps to better spot temporal distortions in STDE.

\subsubsection{Visualization of Temporal Quality Attention}

For the TCT, we aim to enhance the weights of important frames (frames more close to theme) and suppress the weights of unimportant frames (irrelevant frames). In \cref{fig:df}, we visualize the attention matrix $M_\mathrm{QK}$ and final attention weights $\{w_i|_{i=0}^{N-1}\}$ learnt in the TCT. For video 1, the peak attention is at the zoom-in for players, which are the specially important frames for this replay video of the football match, and they result in higher weights for these frames. For video 2, on the contrary, the peak attention comes at the irrelevant frames that photographs on the ground, which are specially irrelevant frames and have lower final weights. These results demonstrates the effectiveness of the proposed TCT.

\subsubsection{Visualization for Correlations with Ground Truth}

To further visualize the result of the proposed DisCoVQA, we show the correlation of DisCoVQA predicted scores and the ground truth labels in \cref{fig:corr}. The x-axis represents ground truth labels (MOS), and the y-axis shows the predicted scores $\hat q$ with respect to MOS. The bright blue line is the reference line when the prediction is the same as MOS. In visualizing these correlations, we add two gray lines as \textbf{unit deviation} for each dataset (the quality prediction is roughly correct if fallen in this line). We find out that the proposed DisCoVQA consistently predicts video scores with a very high correlation with ground truth scores, and only a few videos fall out of the range enclosed by the gray lines. These correlations, together as results shown in \cref{table:cvdqualcommvqckovid}, demonstrate that the proposed DisCoVQA is an accurate quality evaluator.

\subsubsection{Visualization for Success and Failure Cases}

\begin{figure}
    \centering
    \includegraphics[width=\linewidth]{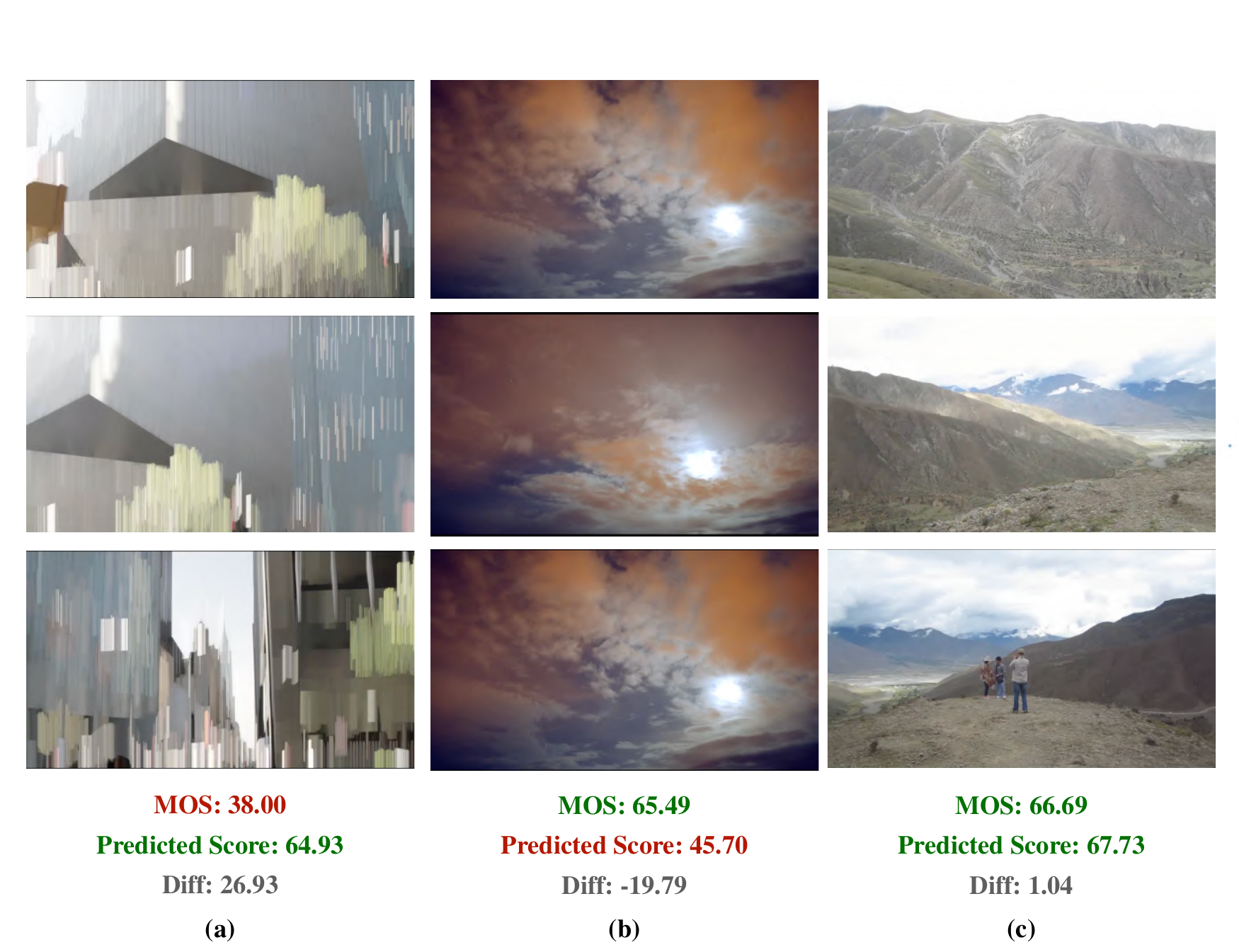}
    \caption{The success and failure cases on dataset LSVQ \cite{pvq} by the proposed DisCoVQA. (a) \textbf{non-natural contents} and (b) \textbf{ambiguity on human annotations} (on a flicker video) are two representative failure cases, where (c) is one of the successful predictions.}
    \label{fig:succfail}
\end{figure}

We also visualize several successful or failed prediction of the proposed model in LSVQ dataset. As the plot in \cref{fig:succfail}(d) shows, the proposed DisCoVQA can give reasonable quality predictions on most videos, including the video \cref{fig:succfail}(c) with complex changing contents across the video. We also visualize two specific failure cases of the model: (a) \textbf{non-natural contents}, which have been discussed above and are also hard situations for several existing methods such as PVQ\cite{pvq}; (b) \textbf{ambiguity of human annotations} (this video contains strong flicker but still gets relatively high MOS), which suggests that the proposed model is especially sensitive on temporal distortions, though for this case the human annotators prefer to give it higher MOS scores. Results of these cases are in line with our analysis for the proposed method.

\subsection{Reliability Analysis for TSF during inference}
\label{sec:rely}

\begin{table}[]
\caption{Reliability analysis for TSF I: the metric results with respect to sample count $M$, with train set LSVQ$_\text{train}$ and test set LIVE-VQC.} \label{table:rapa}
\setlength\tabcolsep{2.4pt}
\renewcommand\arraystretch{1.25}
\footnotesize
\center
\resizebox{\linewidth}{!}{\begin{tabular}{l|cc|cc|cc|cc|cc}
\hline
\textbf{Sample Count} $S_m$       & \multicolumn{2}{|c|}{1}        &  \multicolumn{2}{|c|}{2}  & \multicolumn{2}{|c|}{4}   & \multicolumn{2}{|c|}{8}   & \multicolumn{2}{|c}{$\infty$(40)}            \\ \cline{2-11}
 & SROCC & PLCC           & SROCC & PLCC                      & SROCC & PLCC   & SROCC & PLCC           & SROCC & PLCC                               \\ \hline 

\textbf{DisCoVQA} (proposed) &  0.816 & 0.824 & 0.819 & 0.831 & 0.821 & 0.836 & \textbf{0.823} & \textbf{0.837} & \textbf{0.823} & \textbf{0.837} \\ \hline
\end{tabular}}
\end{table}
\begin{table}
\caption{Reliability analysis for TSF II: the mean standard deviants $\sigma_M$ of different predictions on the same video, with train set LSVQ$_\text{train}$ and test set LIVE-VQC.} \label{table:rastd}
\setlength\tabcolsep{3.9pt}
\renewcommand\arraystretch{1.25}
\footnotesize
\center
\resizebox{\linewidth}{!}{\begin{tabular}{l|c|c|c|c|c|c|c}
\hline
\textbf{Sample Count} $S_m$ & 1&2&4&6&8&16&40  \\ \hline

\textbf{DisCoVQA} (proposed) & 0.0114 & 0.0043 & 0.0025 & 0.0016 & \textbf{0.0010}  & \textbf{0.0006} & \textbf{0.0002} \\ \hline
\end{tabular}}
\end{table}

We have proposed the TSF to significantly reduce the computational complexity of transformer-based modules in VQA. During inference, we random sample features for $S_m=8$ times and get the average quality prediction for these samples. However, as the TSF contains random sampling for frames in each segment, we need to do the following two experiments to confirm its reliability. First, we need to confirm that the TSF will not reduce the overall prediction accuracy of the proposed DisCoVQA. We choose to evaluate on the setting with training set LSVQ$_\text{train}$ and testing set LIVE-VQC, and as the results in \cref{table:rapa} demonstrates, we can obtain nearly the same result as $S_m=8$ compared with infinite samples ($S_m=40$ in practice), proving that the TSF does not harm the overall prediction accuracy. Moreover, we show the mean standard deviants $\sigma_m$ of predictions on different samples (normalized with the overall score range) of the same video to $S_m$ in \cref{table:rastd}, proving that with $S_m\geq8$, the $\sigma_m$ can be negligible (less than 0.001), thus DisCoVQA with TSF can still infer with high stability. Both experiments prove that TSF is reliable during inference.

\section{Conclusion and Future Works}

We have proposed \textbf{DisCoVQA}, a novel and effective method that aims at better modeling both temporal distortions and content-related temporal quality attention via transformer-based architectures. To better capture temporal distortions, we extract multi-level features from a Swin-T backbone network for a better semantic understanding of video actions and compute temporal differences to further spot temporal variations. To model the temporal quality attention toward different importance of frames, we utilize a transformer encoder-decoder structure to 
consider the correlation of frame contents to the overall video theme. We also introduce the temporal sampling on features to boost the training effectiveness and efficiency of this transformer-based temporal regression module. In conclusion, we propose a transformer-based method that better models the temporal relationships in VQA, and the proposed DisCoVQA has reached state-of-the-art performance on several natural VQA datasets and achieved excellent generalization ability among them. 

In the future, we aim at solving several problems not well addressed by current frameworks (as analyzed in failure cases in \cref{fig:succfail}), including the better coverage of non-natural contents, and dealing with ambiguous quality scores. We also notice that several recent methods benefit from extra pre-training, yet they all need labeled datasets. For the next step, we hope to propose a method to include label-free pre-training for VQA that can lead to further improvements on performance.


\ifCLASSOPTIONcaptionsoff
  \newpage
\fi

\bibliographystyle{ieee_fullname}
\bibliography{egbib}







\end{document}